\documentclass{bmvc2k}

\title{Self-Intersection-Aware\\ 3D Human Motion Generation\\ Using an Efficient Human Sphere Proxy}

\addauthor{Pascal Herrmann}{pascal.herrmann@de.bosch.com}{1}
\addauthor{Maarten Bieshaar}{maarten.bieshaar@de.bosch.com}{1}
\addauthor{Dennis Mack}{dennis.mack@de.bosch.com}{1}
\addauthor{Robert Herzog}{paulrobert.herzog@de.bosch.com}{1}
\addauthor{Juergen Gall}{gall@iai.uni-bonn.de}{2,3}

\addinstitution{
Bosch Research
}
\addinstitution{
University of Bonn
}
\addinstitution{
Lamarr Institute for Machine Learning and Artificial Intelligence
}

\runninghead{Herrmann \etal}{Self-Intersection-Aware 3D Human Motion Generation}

\def\etal{\emph{et al}\bmvaOneDot}

\usepackage[capitalize]{cleveref}
\crefname{section}{Sec.}{Secs.}
\Crefname{section}{Section}{Sections}
\Crefname{table}{Table}{Tables}
\crefname{table}{Tab.}{Tabs.}
\usepackage{booktabs}
\usepackage{multirow}
\usepackage{placeins}
\usepackage{scalerel,stackengine,amsmath,bm}

\usepackage{algpseudocode}
\usepackage{algorithm}

\def\vc{{\bm{c}}}

\def\vj{{\bm{j}}}

\def\vp{{\bm{p}}}

\def\vx{{\bm{x}}}

\def\vz{{\bm{z}}}

\def\mB{{\bm{B}}}

\def\mI{{\bm{I}}}

\usepackage{amssymb}
\usepackage{wrapfig}
\usepackage{capt-of}

\begin{document}

\maketitle

\begin{abstract}
Human motion generation has made tremendous progress in recent years, with state-of-the-art approaches surpassing ground truth data in leading evaluation benchmarks. However, visual inspection of the generated motions paints a different picture. Even state-of-the-art approaches generate motions frequently containing self-intersections, i.e., body parts interpenetrating, which are strong artifacts, severely limiting the perceived motion quality.
We introduce a novel loss, which explicitly penalizes self-intersections, to the training of human motion generation methods. We base our loss on a sphere proxy of human geometry, which allows us to calculate a self-intersection loss 98~\% faster and uses 83~\% less memory than comparable methods based on triangular meshes. The loss is agnostic to the specific approach, and we add it to the training of the recent human motion generation methods \textit{human motion diffusion model (MDM)} and \textit{MoMask}. Our extensive experiments show a reduction of self-intersections in generated motions of up to 49~\% while improving other evaluation metrics.
The code is available at \url{https://github.com/boschresearch/humansphereproxy}.
\end{abstract}
\section{Introduction}
\label{sec:introduction}

Generating realistic 3D human motions is an essential task in virtual reality applications~\cite{SLR+24-MHC}, video games~\cite{RKA+22-videogames}, computer animation~\cite{MLT88-3danimation}, and synthetic data generation for deep learning~\cite{VRM+17-synhum}.
Most recent approaches~\cite{TRG+23-MDM, GMJ+24-MoMask, DMG+23-MoFusion} rely on data-driven approaches using transformer-based~\cite{VSP+17-Transformer} diffusion models~\cite{SWM+15-diffusion, HJA20-DDPM, SE20-diffusion} or variational autoencoder (VAE)~\cite{KW-13-VAE, VVK17-VQ-VAE}.
Considering metrics on current evaluation benchmarks~\cite{GZZ+22-humanml3d, PMA16-KIT}, human motion generation has made tremendous progress over the years. However, actually looking at the generated motions is often unsatisfying. Many generated motions contain self-intersections, i.e., body parts interpenetrating, which are strong artifacts that seriously impair the perceived motion quality. This observation highlights two things: 1) Recent human motion generation approaches do not focus enough on the physical plausibility of the generated motions, and 2) current evaluation metrics fail to capture the quality of the generated motions sufficiently.
We aim to avoid self-intersections in generated human motions by explicitly penalizing self-intersections while training human motion generation methods. In the related field of 3D human pose and shape estimation, some approaches~\cite{BKL+16-SMPLify, PCG+19-SMPLX} implement such a loss based on triangular meshes of humans. However, they focus on optimizing the poses of a human for a single motion, while recent human motion generation methods use data batches with many motions. The runtime and memory consumption of these approaches does not scale to this scenario, prohibiting their use.
Instead, we propose to approximate meshes used for training with a set of spheres. This approximation has several advantages.
First, only a few hundred spheres are necessary to approximate a mesh with thousands of triangles. Thus, we can significantly reduce the memory cost of the geometry representation and the number of intersection checks between geometric primitives.
Second, calculating if two spheres intersect is simpler than calculating triangle intersections. Therefore, we can use this sphere proxy to efficiently compute our novel self-intersection loss, which enables us to apply it to recent human motion generation methods.
Our extensive experiments show a significant reduction in generated self-intersections while improving most other metrics.
Our contributions can be summarized as follows:
\begin{itemize}
    \item we introduce a novel self-intersection loss, based on a sphere approximation of human geometry, which reduces the memory cost by 83~\% and the runtime by 98~\% compared to previous approaches, making the self-intersection loss applicable in the first place,
    \item we show that our novel self-intersection loss reduces self-intersections in human motion generation by up to 49~\% while improving other evaluation metrics,
    \item and we introduce a novel voxel-based metric measuring the severity of self-intersections to guide human motion generation towards improving perceived motion quality.
\end{itemize}
\section{Related Work}
\label{sec:related_work}
\begin{figure*}[t!]
    \centering
    \includegraphics[width=1.0\linewidth]{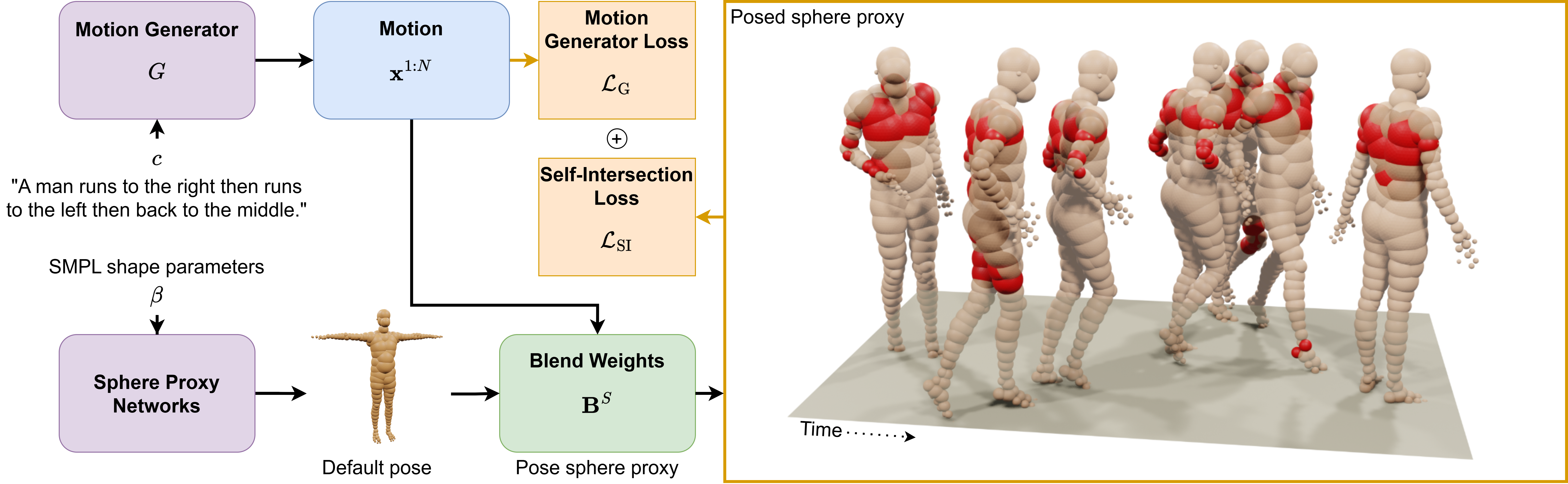}
    \caption{We make human motion generation self-intersection-aware by proposing a novel self-intersection loss. Given a condition embedding $c$, an arbitrary human motion generation method $G$ generates a motion $\vx^{1:N}$. Relying on SMPL shape parameters $\beta$, we obtain our sphere proxy in default pose and employ the blend weight matrix $\mB^S$ to apply the generated poses to the sphere proxy. Our novel self-intersection loss $\mathcal{L}_{\text{SI}}$ is calculated on the sphere intersections of the posed sphere proxy and added to $\mathcal{L}_{\text{G}}$ of $G$. Red indicates intersecting spheres.}
    \label{fig:meth:method_overview}
\end{figure*}
\paragraph{Human Motion Generation}
Recent human motion generation methods use learning approaches on motion capture data~\cite{MGT+19-AMASS, WCL+22-Humanise, BXP+22-BEHAVE, GZZ+22-humanml3d} to either learn the complete manifold of human motions~\cite{RLL+23-MoDi, KTC+24-MAS, ZJG+24-LMM} or condition the generation process on an action class~\cite{PBV21-ACTOR, CSS+22-INR, ZCP+24-motiondiffuse}, audio~\cite{ANB+23-LDA, MDH+24-ConvoFusion}, a motion prefix~\cite{GDG+23-imos, XLW+23-interdiff}, or text~\cite{TRG+23-MDM, PBV22-TEMOS, GMJ+24-MoMask, GZZ+22-humanml3d, CJL+23-MLD, ZZC+23-t2m-gpt, ZGP+23-remodiffuse, RHL-24-CrossDiff, DCW+24-motionlcm}.

MotionDiffuse~\cite{ZCP+24-motiondiffuse} estimates the noise of a noisy motion to iteratively obtain a clean motion~\cite{SWM+15-diffusion, HJA20-DDPM, SE20-diffusion}.
TEMOS~\cite{PBV22-TEMOS} uses a VAE~\cite{KW-13-VAE} by aligning the continuous latent space of a text encoder and a motion encoder, and generates motions with a motion decoder.
Motion latent diffusion~(MLD)~\cite{CJL+23-MLD} applies diffusion models in the latent space of a VAE.
Most recent approaches focus on the controllability of motion generation~\cite{APB+22-TEACH, PLI+24-STMC, XJZ+24-OmniControl, STK+24-priormdm, APB+23-SINC}. In contrast, we focus on the perceived motion quality by avoiding self-intersections in generated motions.
\paragraph{Human Geometry Representations}
Most commonly, human geometries are represented using triangular meshes as in the Skinned Multi-Person Linear~(SMPL)~model~\cite{LMR+15-SMPL, PCG+19-SMPLX, RTB17-MANO, OBB20-STAR}.
Some approaches~\cite{BTG+12-salientpoints, PCG+19-SMPLX, TBS+16-salientpoints} calculate self-intersections for triangular meshes, but they have a high runtime, which can be improved at the cost of memory usage by using space partitioning methods~\cite{K12-bvh, TKH+02-BVH, PCG+19-SMPLX}.
Other representations include signed distance fields (SDF)~\cite{PFS+19-deepSDF, FP06-distancefields}, 3D point clouds~\cite{FSG17-pointcloud}, or occupancy maps~\cite{MON+19-occupancy}.
Finally, human meshes can be approximated by simple geometric primitives. SMPLify~\cite{BKL+16-SMPLify} uses a rough approximation with a set of capsules. Stoll~et~al.~\cite{SHG+11-SoG} use a sum of 3D Gaussians to represent human geometry. DualSDF~\cite{HAS+20-DualSDF} approximates arbitrary geometric shapes using spheres. We extend DualSDF by approximating human meshes with a set of spheres and attaching them to an underlying skeleton.

\paragraph{Physically Grounded Motion Generation}
Enhancing the physical plausibility of generated motions is a growing area of related research.
PhysDiff~\cite{YSI+23-Physdiff} generates physically plausible motions by projecting intermediate noisy motions generated by motion diffusion models into a physics engine.
Several approaches~\cite{TRG+23-MDM, DMG+23-MoFusion} apply geometric losses based on the joints to the training of human motion generation models.
HUMOS~\cite{TTL+24-HUMOS} uses a foot-sliding loss, ground penetration loss, and floating loss based on vertex locations.
3D human pose and shape estimation~\cite{TMH+23-ipman, KAB20-VIBE, SBL+23-trace, QYW+23-PSVT} aims to recover the SMPL~\cite{LMR+15-SMPL} pose and shape parameters given a single image. In this field, losses to enforce physical constraints are frequently used, like a pose prior to penalize unnatural bends of knees and elbows~\cite{BKL+16-SMPLify}, and self-intersection losses~\cite{BKL+16-SMPLify, PCG+19-SMPLX, MSB+22-COAP}.
We follow this line of work by explicitly penalizing self-intersections using a novel loss.
\section{Methodology}
\label{sec:methodology}

We aim to generate physically plausible 3D human motions by explicitly penalizing self-intersections. Our approach is agnostic to the specific human motion generation method, and we explain the general setting in \cref{subsec:meth:HMG}. We base our novel self-intersection loss on a sphere approximation of the human geometry, which we call \textit{sphere proxy} and describe it in \cref{subsec:meth:SP}. \cref{fig:meth:method_overview} shows an overview of our approach. The advantage of the sphere proxy is that self-intersections can be computed significantly more efficiently compared to using triangular meshes. \cref{subsec:meth:SI-loss} describes our novel \textit{self-intersection loss} and how we integrate it into the model training.

\subsection{Human Motion Generation}
\label{subsec:meth:HMG}

Human motion generation aims to generate natural and diverse human motions. Formally, a motion $\vx^{1:N}$ is a discrete temporal sequence of length $N \in \mathbb{N}^+$ of individual poses $\vx^n \in \mathbb{R}^{J \times D}$ with $J \in \mathbb{N}^+$ denoting the number of joints of the underlying skeleton, $D \in \mathbb{N}^+$ denoting the dimension of the pose representation, and $n \in \mathbb{N}^+$ denoting the temporal index. A pose $\vx^n$ can be defined by joint locations, rotations, velocities, or a combination of them. Usually, the motion generation process is conditioned on the embedding of some real-world signal $c \in \mathbb{R}^L$, with $L\in \mathbb{N}^+$ denoting the dimension of the embedding space, like a text description or a 3D point cloud. However, unconditioned generation is also possible. The motion is generated by some generative model $G$ and depends on the specific implementation.

\subsection{Human Sphere Proxy}
\label{subsec:meth:SP}

We follow DualSDF~\cite{HAS+20-DualSDF} to obtain a sphere approximation of human geometries.
DualSDF uses a set of $S \in \mathbb{N}^+$ spheres $\mathbf{S} = \{(\vz^i, r^i) | i = 1,...,S\}$ with centers $\vz^i \in \mathbb{R}^3$ and radii $r^i \in \mathbb{R}$ to approximate a 3D geometry $X$ using SDFs.
Given a point $\vp \in \mathbb{R}^3$, the SDF specifies the distance of that point to the closest surface of the geometry. The sign encodes whether the point lies inside (negative) or outside (positive) the surface.
Given a human mesh $X$, we sample $K \in \mathbb{N}^+$ 3D points $\vp^k, k=1,...,K$, and corresponding SDF values $d_X^k \in \mathbb{R}$.
For a set of spheres $\mathbf{S}$, the value of the SDF at point $\vp$ is defined as the minimum over the SDF values of the individual spheres
\begin{equation}
\label{eq:sdf_sphere_proxy}
    d_{\mathbf{S}} = \min_{1 \leq i \leq S} d_{\text{sphere}}^i, \hspace{1pt} \text{with} \hspace{15pt} d_{\text{sphere}}^i = ||\vp - \vz^i||_2 - r^i.
\end{equation} 
DualSDF implements a neural network to predict the sphere parameters, following the framework of variational autodecoder~(VAD)~\cite{ZLL+19-VAD} by learning a Gaussian latent space. However, a learned latent space would require an optimization process to find the latent vector that corresponds to the sphere proxy of a given mesh. Instead, we use the SMPL~\cite{LMR+15-SMPL} shape parameters $\beta \in \mathbb{R}^U$, $U \in \mathbb{N}^+$ denoting the number of SMPL shape parameters, as input to our neural network, because they have semantic meaning and
they already approximately follow a Gaussian distribution.
We use several loss terms to train the neural network. First of all, we approximate the sampled SDF values of the mesh $X$ using the set of spheres $\mathbf{S}$
\begin{equation}
  \mathcal{L}_{\text{SDF}} = \frac{1}{K} \sum_{k=1}^K
    \begin{cases}
      \max(d_S^k, 0) & d_X^k < 0,\\
      |d_S^k - d_X^k| & d_X^k \geq 0.\\
    \end{cases} 
\end{equation}
For points $\vp^k$ inside the 3D shape $X$, the loss is truncated to zero~\cite{HAS+20-DualSDF, PFS+19-deepSDF}.
Furthermore, we add an emptiness loss.
We want to approximate the surface of the original mesh, so all spheres should be placed within that surface, which we achieve by checking that each sphere contains at least one point $\vp^k$ sampled from inside the mesh.
\begin{equation}
    \mathcal{L}_{\text{emptiness}} = \frac{1}{S} \sum_{i=1}^S
    \begin{cases}
      \max\left(||\vp^{\kappa} - \vz^i|| - r^i, 0\right) & d_X^{\kappa} < 0,\\
      0 & d_X^{\kappa} \geq 0,\\
    \end{cases}   
\end{equation}
where $\kappa = \arg \min_k (||\vp^{k} - \vz^i||)$ is the index of the closest sampled point to sphere center $\vz^i$.
Additionally, the spheres should be distributed equally within the boundaries of the mesh to approximate all body parts with the same level of detail, which we achieve with an intersection loss.
If the distance between the centers $\vz^{i}$ and $\vz^{i'}$ of two spheres $i$ and $i'$ is smaller than the sum of their radii $r^i$ and $r^{i'}$, the spheres are intersecting. The intersection distance is given by
\begin{equation}
    b^{i,i'} = \max(r^i + r^{i'} - ||\vz^i - \vz^{i'}||, 0).
\end{equation}
Minimizing the intersection distance $b^{i,i'}$ pushes the spheres apart from each other, filling the space governed by the boundaries of the mesh. Formally, the loss is given by
\begin{equation}
\label{eq:SP:si-loss}
    \mathcal{L}_{\text{IS}} = \frac{1}{S^2} \sum_{i=1}^S \sum_{i'=i+1}^S b^{i,i'}.
\end{equation}
The overall loss to train the sphere proxy is defined by
\begin{equation}
        \mathcal{L_{\text{SP}}} = \mathcal{L}_{\text{SDF}} + \lambda_{\text{emptiness}} \mathcal{L}_{\text{emptiness}} + \lambda_{\text{IS}} \mathcal{L}_{\text{IS}},
\end{equation}
where $\lambda_{\text{emptiness}}, \lambda_{\text{IS}} \in \mathbb{R}$ are hyperparameters.

We use SMPL meshes in default pose to train the sphere proxy and attach the spheres to the SMPL skeleton to apply a pose $\vx^n$. While learning a posed sphere proxy is also possible, it would require substantially more training resources, as the training data would need to capture various poses sufficiently.
In SMPL, joint locations are regressed, given the vertex locations. Instead, we train a simple neural network to predict them, given SMPL shape parameters using an $L_2$ loss on joint locations.
To attach the spheres to the skeleton, we follow the standard linear blend skinning approach~\cite{MLT88-3danimation}. The sphere's movement is governed by a blend weight matrix $\mB^S \in \mathbb{R}^{S \times J}$. Similarly, the SMPL model also has a blend weight matrix $\mB^X \in \mathbb{R}^{V \times J}$, where $V$ is the number of vertices of the mesh. To obtain $\mB^S$, we calculate the $g \in \mathbb{N}^+$ nearest vertices of the given mesh to the surface of each sphere. The blend weights of a sphere are then simply the mean of the blend weights of the neighboring vertices. Finally, these blend weight matrices are calculated for all meshes in the training data, and subsequently, we take their mean to obtain one blend weight matrix $\mB^S$ for the sphere proxy.
The sphere centers $\vz$ become dependent on a given pose $\vx^n$, $\vz = \vz(\vx^n)$. However, we omit this dependency in the following for ease of notation.

\subsection{Preventing Self-Intersections}
\label{subsec:meth:SI-loss}

We use our sphere proxy to propose a novel self-intersection loss. At some point during the training of every human motion generation method $G$, a human motion $\vx^{1:N}$ is generated. We apply each generated pose to the sphere proxy and calculate which spheres intersect. 
However, it is not ideal to calculate the intersections between all sphere pairs. Neighboring spheres will always intersect, while spheres representing the same body part will never intersect.
We utilize a data-driven approach to determine the spheres that always intersect by recording all sphere intersections given the poses of a human motion dataset. Sphere pairs intersecting in more than $90\%$ of the poses can be seen as body model inaccuracies and are excluded from the loss calculation.
We utilize the blend weight matrix $\mB^S$ to determine the spheres belonging to the same body part by assigning each sphere to the joint for which the blend weight is the biggest. Sphere pairs assigned to the same joint are excluded from the loss calculation. We denote the remaining set of sphere pairs checked in the self-intersection loss as $\mathbf{W}$ and define the self-intersection loss using the intersection distance between sphere pairs
\begin{equation}
    \mathcal{L}_{\text{SI}} = \frac{1}{N} \sum_{n=1}^N \sum_{(i,i') \in \mathbf{W}} \left(b^{i,i'}\right)^2.
\end{equation}
Note, that $b^{i,i'}$ is dependent on a given pose $\vx^n$.
The overall training loss for $G$ becomes
\begin{equation}
    \mathcal{L} = \mathcal{L}_{\text{G}} + \lambda_{\text{SI}} \mathcal{L}_{\text{SI}},
\end{equation}
where $\mathcal{L}_{\text{G}}$ is the training loss of $G$ and $\lambda_{\text{SI}} \in \mathbb{R}$ is a hyperparameter.
\section{Experiments}
\label{sec:experiments}
We evaluate our methods on the text-to-motion task on the datasets HumanML3D~\cite{GZZ+22-humanml3d} and KIT-ML~\cite{PMA16-KIT}.
We use the motion representation proposed by Guo~\etal~\cite{GZZ+22-humanml3d} but also recover SMPL~\cite{LMR+15-SMPL} joint rotations to pose our sphere proxy - details can be found in the supplementary material. KIT-ML follows a different skeletal structure than HumanML3D, but TEMOS~\cite{PBV22-TEMOS} provides correspondences between the KIT-ML and SMPL joints while we interpolate missing joints.
HumanML3D and KIT-ML represent motions using the skeleton of one target motion. We apply SMPLify~\cite{BKL+16-SMPLify} to each target motion to get one set of SMPL shape parameters for each dataset to obtain our sphere proxy.
We use the metrics \textit{R Precision}, \textit{Fréchet Inception Distance} (\textit{FID}), \textit{Multimodal distance}, \textit{Diversity}, and \textit{MultiModality} to evaluate our approach as commonly used in the literature~\cite{GZZ+22-humanml3d}.
In addition, we propose a novel metric, \textit{Self-Intersect (SI)}, to measure the severity of self-intersections in the generated motions. \textit{SI} approximates the mean self-intersection volume by generating a mesh for every generated pose, scaling each mesh to fit into a sphere with radius $1$~m, subdividing the space within the sphere into voxels with edge length $v=0.06$~cm, and determining if the voxel lies within a region of self-intersection. Values are reported in cubic centimeters. Full details on \textit{SI} can be found in the supplementary material.
\subsection{Motion Generation Methods}
\label{subsec:ex:baseline}
\begin{table}[t]
    \centering
    \resizebox{1.0\linewidth}{!}{
    \begin{tabular}{lllccccccc}
    \toprule
    \multirow{2}{2.cm}{\centering Dataset} &
    \multirow{2}{2.cm}{\centering Method} &
    \multirow{2}{2.cm}{\centering R Precision top 1$\uparrow$} & 
    \multirow{2}{2.cm}{\centering R Precision top 2$\uparrow$} & 
    \multirow{2}{2.cm}{\centering R Precision top 3$\uparrow$} & 
    \multirow{2}{1.5cm}{\centering FID$\downarrow$} &
    \multirow{2}{2.5cm}{\centering Multimodal Dist$\downarrow$} &
    \multirow{2}{2cm}{\centering Diversity$\rightarrow$} &
    \multirow{2}{2cm}{\centering MultiModality$\uparrow$} &
     \multirow{2}{1.5cm}{\centering SI$\downarrow$} \\
    \\
    
    \midrule
        \multirow{8}{*}{HumanML3D} & Real & $0.511^{\pm 0.003}$ & $0.703^{\pm 0.003}$ & $0.797^{\pm 0.002}$ & $0.002^{\pm 0.000}$ & $2.974^{\pm 0.008}$ & $9.503^{\pm 0.065}$ & - & $447^{\pm 0}$ \\
        
    \cline{2-10}
        \rule{0pt}{2.5ex}& MLD~\cite{CJL+23-MLD} & $0.481^{\pm 0.003}$ & $0.673^{\pm 0.003}$ & $0.772^{\pm 0.002}$ & $0.473^{\pm 0.013}$ & $3.196^{\pm 0.010}$ & $9.724^{\pm 0.082}$ & $2.413^{\pm 0.079}$ & - \\
        & MotionDiffuse~\cite{ZCP+24-motiondiffuse} & $0.491^{\pm 0.001}$ & $0.681^{\pm 0.001}$ & $0.782^{\pm 0.001}$ & $0.630^{\pm 0.001}$ & $3.113^{\pm 0.001}$ & $9.410^{\pm 0.049}$ & $1.553^{\pm 0.042}$ & - \\
        & ReMoDiffuse~\cite{ZGP+23-remodiffuse} & $0.510^{\pm 0.005}$ & $0.698^{\pm 0.006}$ & $0.795^{\pm 0.004}$ & $0.103^{\pm 0.004}$ & $2.974^{\pm 0.016}$ & $9.018^{\pm 0.075}$ & $1.795^{\pm 0.043}$ & - \\
        & MDM~\cite{TRG+23-MDM} & $0.418^{\pm 0.005}$ & $0.604^{\pm 0.005}$ & $0.707^{\pm 0.004}$ & $0.489^{\pm 0.025}$ & $3.631^{\pm 0.023}$ & $\mathbf{9.449^{\pm 0.066}}$ & $\mathbf{2.973^{\pm 0.111}}$ & $619^{\pm 11}$ \\
        & MoMask~\cite{GMJ+24-MoMask} & \underline{$0.521^{\pm 0.002}$} & \underline{$0.713^{\pm 0.002}$} & \underline{$0.807^{\pm 0.002}$} & $\mathbf{0.045^{\pm 0.002}}$ & \underline{$2.958^{\pm 0.008}$} & $9.624^{\pm 0.080}$ & $1.241^{\pm 0.040}$ & \underline{$316^{\pm 02}$} \\
    \cline{2-10}
        \rule{0pt}{2.5ex}& SIA-MDM (Ours) & $0.435^{\pm 0.005}$ & $0.628^{\pm 0.006}$ & $0.731^{\pm 0.006}$ & $0.265^{\pm 0.024}$ & $3.462^{\pm 0.026}$ & \underline{$9.568^{\pm 0.086}$} & \underline{$2.893^{\pm 0.075}$} & $382^{\pm 09}$ \\
        & SIA-MoMask (Ours) & $\mathbf{0.525^{\pm 0.003}}$ & $\mathbf{0.717^{\pm 0.003}}$ & $\mathbf{0.813^{\pm 0.002}}$ & \underline{$0.068^{\pm 0.002}$} & $\mathbf{2.933^{\pm 0.006}}$ & $9.691^{\pm 0.092}$ & $1.198^{\pm 0.041}$ & $\mathbf{290^{\pm 02}}$ \\

    \midrule
        \multirow{8}{*}{KIT-ML} & Real & $0.424^{\pm 0.005}$ & $0.649^{\pm 0.006}$ & $0.799^{\pm 0.006}$ & $0.031^{\pm 0.004}$ & $2.788^{\pm 0.012}$ & $11.080^{\pm 0.097}$ & - & $778^{\pm 0}$ \\

    \cline{2-10}

        \rule{0pt}{2.5ex}& MLD~\cite{CJL+23-MLD} & $0.390^{\pm 0.008}$ & $0.609^{\pm 0.008}$ & $0.734^{\pm 0.007}$ & $0.404^{\pm 0.027}$ & $3.204^{\pm 0.027}$ & $10.800^{\pm 0.117}$ & \underline{$2.192^{\pm 0.071}$} & - \\
        & MotionDiffuse~\cite{ZCP+24-motiondiffuse} & $0.417^{\pm 0.004}$ & $0.621^{\pm 0.004}$ & $0.739^{\pm 0.004}$ & $1.954^{\pm 0.062}$ & $2.958^{\pm 0.005}$ & $\mathbf{11.100^{\pm 0.143}}$ & $0.730^{\pm 0.013}$ & - \\
        & ReMoDiffuse~\cite{ZGP+23-remodiffuse} & $0.427^{\pm 0.014}$ & \underline{$0.641^{\pm 0.004}$} & $0.765^{\pm 0.055}$ & $\mathbf{0.155^{\pm 0.006}}$ & $2.814^{\pm 0.012}$ & $10.800^{\pm 0.105}$ & $1.239^{\pm 0.028}$ & - \\
        & MDM~\cite{TRG+23-MDM} & $0.404^{\pm 0.005}$ & $0.607^{\pm 0.004}$ & $0.731^{\pm 0.004}$ & $0.513^{\pm 0.046}$ & $3.096^{\pm 0.024}$ & $10.732^{\pm 0.103}$ & $1.806^{\pm 0.176}$ & $597^{\pm 07}$ \\
        & MoMask~\cite{GMJ+24-MoMask} & \underline{$0.433^{\pm 0.007}$} & $\mathbf{0.656^{\pm 0.005}}$ & $\mathbf{0.781^{\pm 0.005}}$ & \underline{$0.204^{\pm 0.011}$} & \underline{$2.779^{\pm 0.022}$} & $10.780^{\pm 0.080}$ & $1.131^{\pm 0.043}$ & $930^{\pm 07}$ \\
    \cline{2-10}
        \rule{0pt}{2.5ex}& SIA-MDM (Ours) & $0.416^{\pm 0.005}$ & $0.635^{\pm 0.006}$ & $0.755^{\pm 0.006}$ & $0.321^{\pm 0.021}$ & $2.981^{\pm 0.024}$ & \underline{$10.922^{\pm 0.107}$} & $\mathbf{2.234^{\pm 0.080}}$ & $\mathbf{441^{\pm 05}}$ \\
        & SIA-MoMask (Ours) & $\mathbf{0.437^{\pm 0.005}}$ & $\mathbf{0.656^{\pm 0.006}}$ & \underline{$0.776^{\pm 0.005}$} & $0.316^{\pm 0.017}$ & $\mathbf{2.722^{\pm 0.017}}$ & $10.709^{\pm 0.120}$ & $1.111^{\pm 0.034}$ & \underline{$472^{\pm 03}$} \\

    \bottomrule
    \end{tabular}
        }
    \caption{Text-to-motion results on HumanML3D~\cite{GZZ+22-humanml3d} and KIT-ML~\cite{PMA16-KIT}. All experiments are repeated for 20 random seeds. $\pm$ indicates the 95\% confidence interval. \textbf{Bold} indicates the best result, while \underline{underscore} indicates the second best. $\rightarrow$ indicates that closer to 'Real' is better. Our self-intersection-aware methods significantly improve the respective baselines in most metrics.}
    \label{tab:exp:t2m}
\end{table}

We integrate our novel self-intersection loss into the training of the recent human motion generation methods \textit{human motion diffusion model (MDM)}~\cite{TRG+23-MDM} and \textit{MoMask}~\cite{GMJ+24-MoMask}, which we briefly explain in the following. We call our modified versions \textit{SIA-MDM} and \textit{SIA-MoMask}, with SIA being short for self-intersection-aware.

\textbf{MDM} follows the diffusion model framework by learning a reverse diffusion process. Given a noisy motion $\vx_t^{1:N}$ at noise step $t \in \mathbb{N}^+$, MDM implements $G$ with a Transformer~\cite{VSP+17-Transformer} encoder-only architecture and is trained to predict the clean motion $\hat{\vx}_0^{1:N}$, given the condition embedding $c$ and the noise step $t$.
During sampling, $\hat{\vx}_0^{1:N}$ is passed through the forward diffusion process to obtain $\vx_{t-1}^{1:N}$ and this procedure is iterated until $\vx_0^{1:N}$ is obtained. To generate a motion, $\vx_T^{1:N} \sim \mathcal{N}(0, \mI)$ is sampled from a Gaussian distribution and iteratively denoised, where $T$ denotes the maximal noise step.

\textbf{MoMask} follows the vector quantized VAE (VQ-VAE)~\cite{VVK17-VQ-VAE} framework and implements $G$ with three components. First, a residual VQ-VAE is trained, which encodes a motion $\vx^{1:N}$ using a hierarchy of quantization layers of discrete motion tokens corresponding to entries of a learned codebook. Second, a masked transformer is trained to generate motion tokens of the base layer of the quantization hierarchy given the condition embedding $c$. Third, a residual transformer is trained to generate the motion tokens of the remaining quantization layers given $c$. During sampling, the motion tokens of the quantization hierarchy are progressively generated using the masked and residual transformer given $c$. Subsequently, the motion tokens are decoded using the decoder of the residual VQ-VAE.
    
\subsection{Implementation Details}
\label{subsec:ex:implementation}
\paragraph{Sphere Proxy}
The sphere proxy is based on the gender-neutral SMPL-H~\cite{RTB17-MANO} model without the hand joints and SMPL shape parameters of dimension $U=10$ in a range between $-5.0$ and $5.0$, following Guo~et~al.~\cite{GZZ+22-humanml3d}. We randomly sample 8,000 sets of SMPL shape parameters and use the corresponding joint locations to train the joint regressor. It is trained for 300 epochs with an initial learning rate of 1e-4, which is decayed by 0.1 every 100 epochs.
We randomly sample 800 sets of SMPL shape parameters to train the sphere regressor. For each corresponding mesh, we sample $K=750,000$ points with associated SDF values, of which 250,000 points are sampled in a sphere around the mesh, and 500,000 points are sampled closely to the surface.
The sphere regressor is trained for 2,800 epochs with an initial learning rate of 5e-4, which is decayed by 0.5 every 700 epochs. Each batch contains 16,384 SDF samples per mesh, of which 10\% are sphere samples, and of the remaining points, 50\% correspond to hands and feet, as these body parts have more details. We empirically set $\lambda_{\text{emptiness}} = 10$ and $\lambda_{\text{IS}} = 0.1$.
Both models are trained using the Adam~\cite{KB15-ADAM} optimizer with a batch size of $64$. The architectures of both models are shown in the supplementary material.
We use the mean of the blend weights of the $g = 8$ nearest neighbor vertices and only keep the four most significant values following SMPL~\cite{LMR+15-SMPL}. Our sphere proxy contains $S=192$ spheres unless stated otherwise. For the intersection reduction, we use the poses of the training split of the HumanML3D dataset.
\begin{figure*}
    \centering
\includegraphics[width=1.0\linewidth]{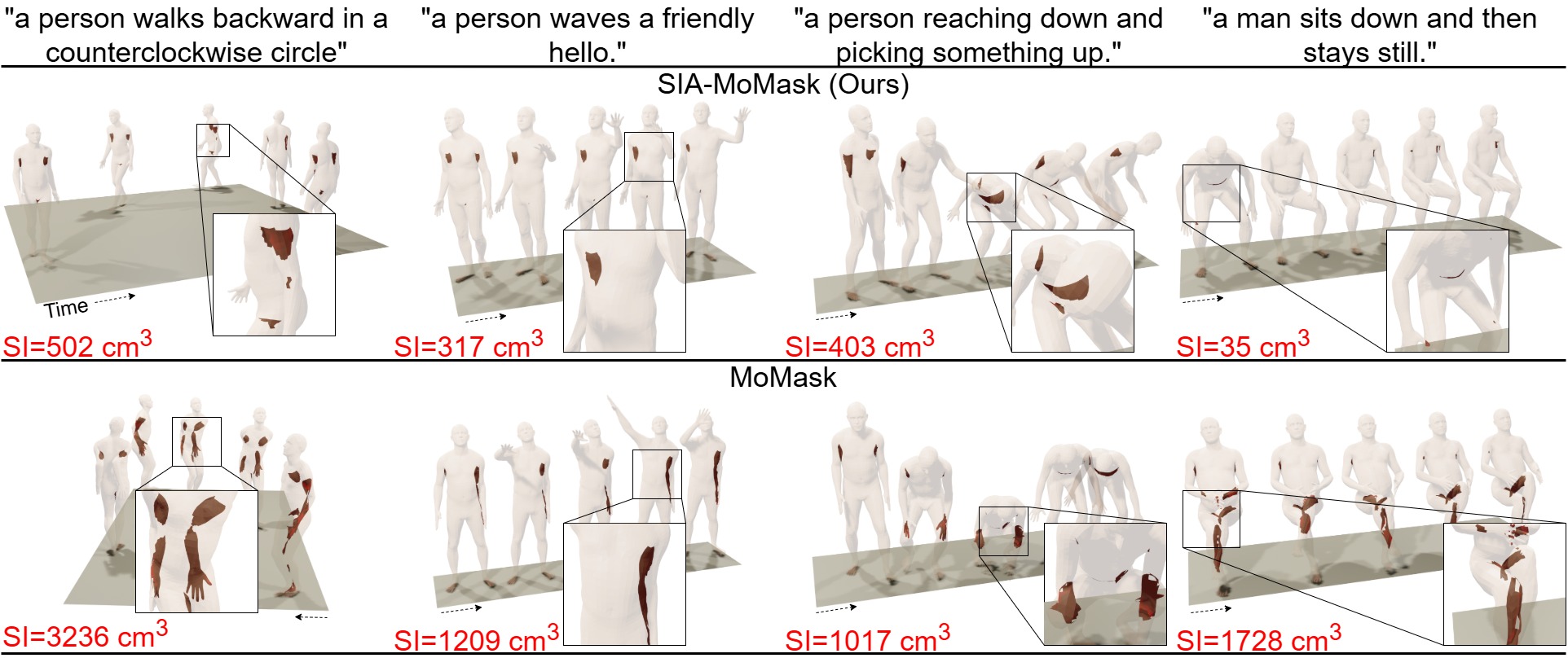}
    \caption{\textbf{Visual Comparison} between motions generated by SIA-MoMask (Ours) and MoMask~\cite{GMJ+24-MoMask} given text prompts from the HumanML3D~\cite{GZZ+22-humanml3d} test set. Red patches indicate self-intersections. \textcolor{red}{SI} indicates the mean self-intersection volume. SIA-MoMask generates significantly fewer self-intersections while semantically following the textual description.}
    \label{fig:exp:qualitative}
\end{figure*}
\paragraph{Motion Generation Methods}
We run our experiments on SIA-MDM and SIA-MoMask using the parameters given in the available code of MDM and MoMask, respectively. SIA-MDM is trained for 600,000 steps with $\lambda_{\text{IS}} = 0.01$ on HumanML3D and for 200,000 steps with $\lambda_{\text{SI}} = 0.0001$ on KIT-ML. SIA-MoMask integrates our novel loss into the residual VQ-VAE training with $\lambda_{\text{SI}} = 0.01$ for HumanML3D, and into the training of all components with $\lambda_{\text{SI}} = 0.000001$ for KIT-ML. Text conditions are embedded using CLIP~\cite{RKH+21-CLIP}.

\subsection{Results}
\label{subsec:ex:results}

We compare the results obtained with SIA-MDM and SIA-MoMask to the results of MDM\footnote{The evaluation script of the original publication of MDM contained errors. All results are obtained following bug fixes detailed at \url{https://github.com/GuyTevet/motion-diffusion-model/issues/182}} and MoMask in \cref{tab:exp:t2m}.
Both self-intersection-aware methods generate significantly fewer self-intersections than the respective baselines, emphasizing the effectiveness and generality of our novel loss. Additionally, most other metrics improve, highlighting the benefit of focusing on the physical plausibility of generated motions.
Furthermore, we note the substantial amount of self-intersections in the ground truth data, the origin of which we discuss in the supplementary material. Our novel self-intersection loss facilitates the generation of motions containing few self-intersections, even if they are present in the ground truth data.
However, we acknowledge that MDM and MoMask also generate fewer self-intersections than the ground truth motions on KIT-ML and HumanML3D, respectively, which we attribute to the desired diversity of motion generation, thus deviating from the ground truth. Nevertheless, focusing solely on optimizing FID, often associated with motion quality, does not result in the most realistic motions. With FID approaching the ground truth data, the amount of generated self-intersections is also expected to approach the ground truth self-intersections, highlighting the limitation of the FID metric. Therefore, evaluating FID together with SI yields a better judgment of motion quality.
\cref{fig:exp:qualitative} and the supplementary material show motions generated by MoMask and SIA-MoMask given text prompts of the HumanML3D test set, which confirms the improved motion quality compared to the baseline models.

\subsection{Ablation Studies}
\label{subsec:ex:ablations}

We ablate design choices on the HumanML3D~\cite{GZZ+22-humanml3d} dataset using SIA-MDM and focus on the metrics \textit{FID}, \textit{MultiModality}, and \textit{SI}. All results are shown in \cref{tab:exp:abl}.

\begin{table}[ht]
\begin{minipage}[b]{0.56\linewidth}
\centering
\resizebox{1.0\linewidth}{!}{
    \begin{tabular}{lccc}
    \toprule
    \multirow{2}{2.cm}{\centering Method} &
    \multirow{2}{1.5cm}{\centering FID$\downarrow$} &
    \multirow{2}{2cm}{\centering MultiModality$\uparrow$} &
    \multirow{2}{2cm}{\centering SI$\downarrow$} \\
    \\
    \midrule
        No intersection reduction & $0.841^{\pm 0.058}$ & $3.083^{\pm 0.067}$ & $715^{\pm 12}$ \\
    \midrule
    \midrule
        128 spheres & $0.411^{\pm 0.056}$ & $2.711^{\pm 0.059}$ & $306^{\pm 11}$ \\
        %192 spheres & $0.265^{\pm 0.024}$ & $2.893^{\pm 0.075}$ & $381.595^{\pm 08.768}$ \\
        256 spheres & $0.301^{\pm 0.038}$ & $2.754^{\pm 0.061}$ & $331^{\pm 05}$ \\
    \midrule
    \midrule
        SIA-MDM (Ours) & $0.265^{\pm 0.024}$ & $2.893^{\pm 0.075}$ & $382^{\pm 09}$ \\
        MDM~\cite{TRG+23-MDM} & $0.489^{\pm 0.025}$ & $2.973^{\pm 0.111}$ & $619^{\pm 11}$ \\
    \bottomrule

    \end{tabular}
    }
    \caption{Ablation studies on HumanML3D~\cite{GZZ+22-humanml3d} using SIA-MDM. All experiments are repeated for 20 random seeds. $\pm$ indicates the 95\% confidence interval.}
    \label{tab:exp:abl}
\end{minipage}\hfill
\begin{minipage}[b]{0.4\linewidth}
\centering
\includegraphics[width=0.7\textwidth]{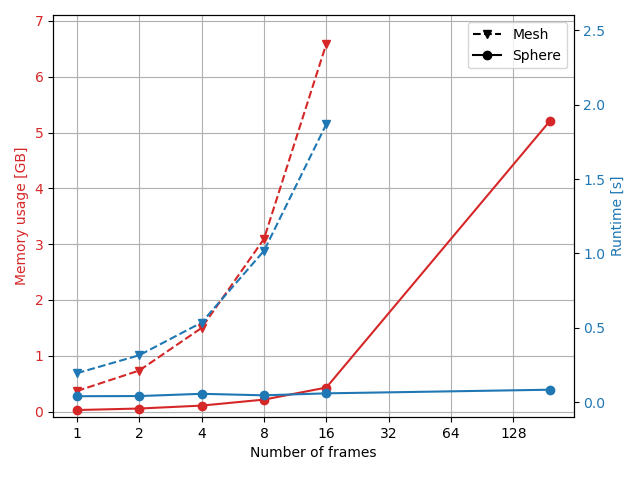}
  \captionof{figure}{Comparison of the computational efficiency of the self-intersection loss between meshes and the sphere proxy.}
  \label{fig:exp:mem-time-usage}
\end{minipage}
\end{table}

\paragraph{Influence of the self-intersection method}
In \cref{fig:exp:mem-time-usage}, we compare the memory usage and runtime of our novel loss to the implementation of SMPLify-X~\cite{PCG+19-SMPLX} using meshes by applying each loss to 1, 2, 4, 8, and 16 frames of each motion.
Memory usage and runtime increase dramatically with the number of frames for the mesh-based approach, prohibiting its use in human motion generation. In contrast, the runtime of the sphere proxy stays almost constant while the memory consumption is reasonable, even when using all frames.
\paragraph{Influence of intersection reduction}
We use all sphere pairs to calculate our self-intersection loss and compare it to our sphere pair reduction.
Using all sphere pairs results in worse performance than using the self-intersection reduction and no self-intersection loss at all. We believe the additional self-intersections during the loss computation result in a strong but unclear gradient signal hindering model optimization. Additionally, penalizing realistic poses that contain self-intersections due to body model inaccuracies might impair model performance. This result highlights the importance of intersection reduction. 
\paragraph{Influence of number of spheres}
We compare our sphere proxy to versions using $128$ and $256$ spheres. All versions improve MDM in \textit{FID} and \textit{SI}. However, $192$ spheres yield the best trade-off between \textit{FID} and \textit{SI}; hence, we use this version in our experiments.
\paragraph{Evaluation of the sphere proxy}
We randomly sample 150 sets of SMPL shape parameters and use the corresponding meshes as a test set for our sphere proxy. We compute the SDF value of each mesh vertex for the corresponding sphere proxy and take the mean of the absolute SDF values for all vertices. The mean distance per vertex is approximately $0.6$~cm, verifying the sphere proxy as a valid approximation of the surface of human meshes. The supplementary material provides a thorough analysis of the sphere proxy.
\section{Conclusion}
\label{sec:conclusion_future_work}
This paper investigates the problem of self-intersections in human motion generation, and we show that even state-of-the-art human motion generation approaches suffer from them, limiting the perceived motion quality.
To mitigate this problem, we propose a sphere proxy of human geometry and show computational superiority in calculating self-intersections regarding runtime and memory usage compared to triangular meshes. Integrating our novel self-intersection loss into the training of MDM~\cite{TRG+23-MDM} and MoMask~\cite{GMJ+24-MoMask} significantly reduces self-intersections in generated motions while improving other evaluation metrics and the perceived motion quality, as we show with visual examples.
Future research can use our sphere proxy to model contact between humans interacting with a scene or other humans.
Finally, we acknowledge that the sphere proxy is only an approximation of human geometry, and some unrealistic self-intersections are still generated. Future research could further improve the sphere proxy, e.g., by incorporating hand and finger motions.

\paragraph{Acknowledgement} Juergen Gall has been supported by the ERC Consolidator Grant FORHUE (101044724).

\bibliography{egbib}
\appendix
\clearpage
\setcounter{page}{1}

   \newpage
        \begin{centering}
        \Large
        \textcolor{bmv@sectioncolor}{\textbf{Self-Intersection-Aware
3D Human Motion Generation
Using an Efficient Human Sphere Proxy}}\\
        \vspace{0.5em}Supplementary Material \\
        \vspace{1.0em}
        \end{centering}

In this supplementary material, we provide additional information and results. \cref{sec:supp:model_architecture} explains the model architectures of the neural networks used in the sphere proxy. \cref{sec:supp:gt} further analyzes the number of self-intersections in the ground truth data. \cref{sec:supp:metric} shows details for the calculation of our novel self-intersection metric \textit{SI}. \cref{sec:supp:lambda_intersect} investigates the impact of the hyperparameter $\lambda_{\text{SI}}$ on SIA-MDM and SIA-MoMask in-depth. \cref{sec:supp:hml-to-smpl} details a novel method to recover SMPL~\cite{LMR+15-SMPL} pose parameters directly from the given HumanML3D~\cite{GZZ+22-humanml3d} pose representation. \cref{sec:supp:abblation} provides additional results on the ablation studies. \cref{sec:supp:sphere_proxy} thoroughly analyzes the sphere proxy. Finally, \cref{sec:supp:qualitative} provides additional qualitative results.

\section{Model Architectures}
\label{sec:supp:model_architecture}

\begin{figure}[ht]
\centering
\subfigure[Architecture of the sphere regressor.]{
    \includegraphics[width=0.47\textwidth]{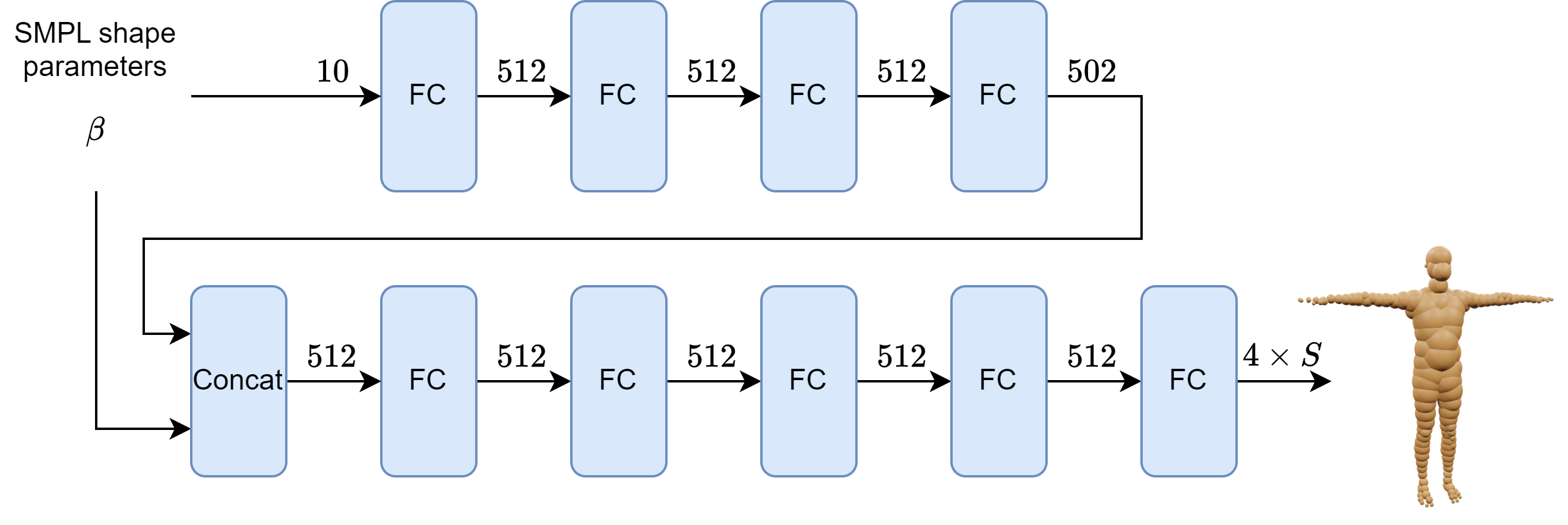}
    \label{fig:supp:sphereregressor}
}
\subfigure[Architecture of the joint regressor.]{
    \includegraphics[width=0.47\textwidth]{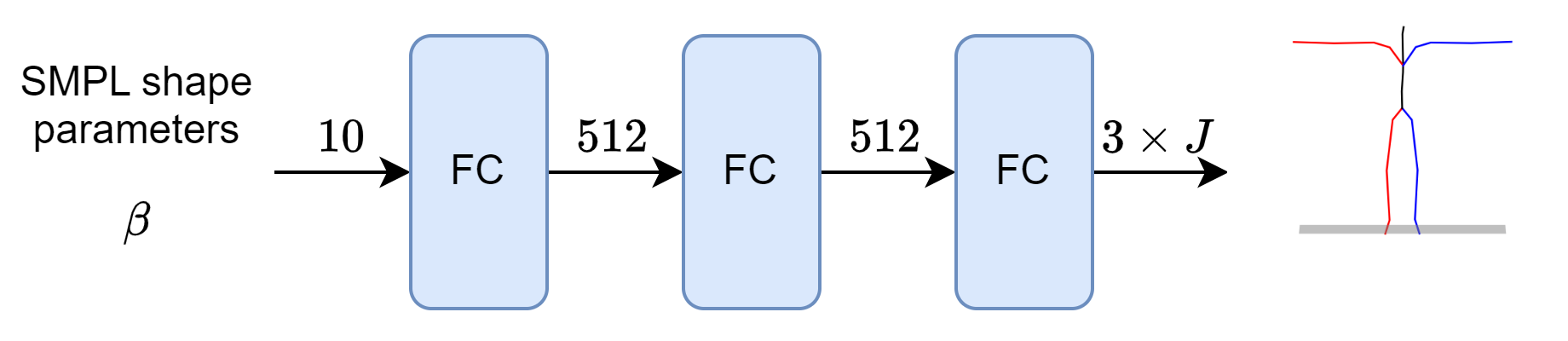}
    \label{fig:supp:jointregressor}
}      
\caption{Neural network architectures of the sphere proxy. FC denotes fully connected layers. Numbers on top of arrows represent the dimensions of the features. $S$ stands for the number of spheres used in the sphere proxy. $J$ denotes the number of joints of the underlying skeleton, whereas $\beta$ denotes the SMPL shape parameters.}
\label{fig:supp:neural_network_architectures}
\end{figure}
\paragraph{Sphere Regressor}
The architecture of the sphere regressor follows the architecture used in DualSDF~\cite{HAS+20-DualSDF} and is shown in \cref{fig:supp:sphereregressor}. Every fully connected layer is followed by weight normalization~\cite{SK16-weightnormalization} and a ReLU activation function. The latent dimension is $512$. As input, we use the shape parameters of the gender-neutral SMPL-H~\cite{LMR+15-SMPL, RTB17-MANO} model of dimension $U=10$ instead of learning a Gaussian latent space as done by DualSDF. A learned latent space would require an optimization procedure to obtain the latent vector representing the sphere proxy for a given mesh. We use the SMPL shape parameters as our latent representation due to their semantic meaning, i.e., a direct correspondence between a shape parameter and a mesh, and because the SMPL shape parameters approximately follow a Gaussian distribution. The output of the sphere regressor is the radii and center locations of the $S$ spheres.
\paragraph{Joint Regressor}
The architecture of the joint regressor is a simple feed-forward network with latent dimension $512$ and is shown in \cref{fig:supp:jointregressor}. Every fully connected layer is followed by a ReLU activation function. As input, we use the shape parameters of the gender-neutral SMPL-H~\cite{LMR+15-SMPL, RTB17-MANO} model of dimension $U=10$. The neural network output is the joint locations of the gender-neutral SMPL-H~\cite{LMR+15-SMPL, RTB17-MANO} model without the hand joints, following Guo~et~al.~\cite{GZZ+22-humanml3d}, resulting in $J=22$ joints.

\section{Ground Truth Self-Intersections}
\label{sec:supp:gt}

\begin{table}[ht]
    \centering
    \begin{tabular}{llc}
    \toprule
    \multirow{2}{2.cm}{\centering Dataset} &
    \multirow{2}{2.cm}{\centering Method} &    
    \multirow{2}{2.cm}{\centering SI $\downarrow$} \\
    \\
    \midrule
         \multirow{6}{2.cm}{\centering HumanML3D} & GT$_{\text{AMASS}}$ & $237^{\pm 00}$ \\
         & GT & $447^{\pm 00}$ \\
         & MDM & $619^{\pm 11}$ \\
         & MoMask & $316^{\pm 02}$\\
         & SIA-MDM (Ours) & $382^{\pm 09}$ \\
         & SIA-MoMask (Ours) & $290^{\pm 02}$ \\
    \bottomrule
    \end{tabular}
    \caption{Self-intersection results on HumanML3D for SIA-MoMask (Ours), SIA-MDM (Ours), MoMask~\cite{GMJ+24-MoMask}, MDM~\cite{TRG+23-MDM}, and the ground truth data. All experiments are repeated for 20 random seeds. $\pm$ indicates the 95\% confidence interval.}
    \label{tab:supp:SI-results}
\end{table}

HumanML3D processes motions of the datasets AMASS~\cite{MGT+19-AMASS} and HumanAct12~\cite{GZW+20-Humanact12} and provides text annotations for them. Part of the processing is the projection of all motions to one target skeleton. To investigate the influence of this projection on self-intersections present in the motions, we calculate \textit{SI} for the original motions of the AMASS dataset, denoted as GT$_{\text{AMASS}}$, and compare it to the processed motions of HumanML3D, denoted as GT.
Projecting all motions to one target skeleton introduces new self-intersections to the motion data. This increase is reasonable because people of different sizes and proportions move differently. This difference in movement is ignored when projecting all motions to one skeleton.
In addition, it is worth mentioning that the motions in AMASS also contain self-intersections, even though they are captured using motion capture systems, highlighting the limitations of the SMPL model. \cref{fig:supp:more_qual_res} shows example motions of the HumanML3D test set. Most self-intersections in the ground truth data are near the armpits and between the legs. In the real world, these body parts would deform under close contact. However, SMPL does not model these deformations, resulting in self-intersections.
Our novel self-intersection loss alleviates the problem of self-intersections in ground truth data. SIA-MDM generates fewer self-intersections as contained in the HumanML3D test set. However, avoiding all self-intersections would result in stiff and unrealistic motions, as some self-intersections are due to SMPL model inaccuracies while still being realistic.

\section{Self-Intersection Metric}
\label{sec:supp:metric}

We describe only briefly in \cref{sec:experiments} how to calculate our novel metric \textit{SI}. Here, we provide full details on the calculation of the metric.
For each pose of the generated motions, we pose SMPL meshes, place them at the origin, and scale them to fit into a sphere with a radius of $1$~m to make the metric unaffected by the mesh size. Next, we subdivide the space within the sphere into voxels with edge length $v=0.06$~cm for a good trade-off between accuracy and computational effort. For each voxel, we cast rays in random directions and record the number and direction of triangle collisions of that ray. We denote $\chi_{m}^{n} \in \mathbb{N}^+$ as the number of surplus triangle back collisions for voxel $m$ of pose $n$. For $\chi_{m}^{n} = 0$, the voxel is outside the mesh, for $\chi_{m}^{n} = 1$, the voxel is inside the mesh but without self-intersections, and for $\chi_{m}^{n} \geq 2$, the voxel is at a self-intersection region. The approximated self-intersection volume of a single mesh is the sum of all voxels with $\chi_{m}^{n} \geq 2$, weighted by $\chi_{m}^{n}$.
We take the mean of the individual self-intersection volumes for the test set containing $N$ poses. Mathematically, this is denoted as
\begin{equation}
    \text{SI} = \frac{1}{N} \sum_{n=1}^{N} \sum_{m=1}^{M} v^3 \chi_{m}^{n}[\chi_{m}^{n} \geq 2],
\end{equation}
where $M \in \mathbb{N}^+$ is the number of voxels in the unit sphere. $[\chi_{m}^{n} \geq 2] = 1$ for $\chi_{m}^{n} \geq 2$ and $[\chi_{m}^{n} \geq 2] = 0$ for $\chi_{m}^{n} < 2$. Values are reported in cubic centimeters.
\cref{alg:supp:SI-Metric} summarizes the calculation of our novel self-intersection metric.

\begin{algorithm}
\caption{Calculate Self-Intersect Metric}
\begin{algorithmic}
\State $SI \gets [\ ]$
\For{each pose $n$ in the generated motions}
    \State $V_{SI} \gets 0$
    \For{each voxel $m$}
        \State $\chi_{m}^{n} \gets 0$
        \State Cast rays in random directions
        \State Record ray-triangle collisions
        \For{each triangle collision}
            \If{triangle hit from the back}
                \State $\chi_{m}^{n} \gets \chi_{m}^{n} + 1$
            \Else
                \State $\chi_{m}^{n} \gets \chi_{m}^{n} - 1$
            \EndIf
        \EndFor
        \If{$\chi_{m}^{n} \geq 2$}
            \State $V_{SI} \gets V_{SI} + v^3 \chi_{m}^{n}$
        \EndIf
    \EndFor
    \State Append $V_{SI}$ to $SI$
\EndFor
\State Calculate the mean of $SI$
\end{algorithmic}
\label{alg:supp:SI-Metric}
\end{algorithm}

\section{Hyperparameter Tuning}
\label{sec:supp:lambda_intersect}

\paragraph{SIA-MDM}
We investigate the influence of the hyperparameter $\lambda_{\text{SI}}$ on the model performance of SIA-MDM. For this purpose, we train different versions of SIA-MDM using varying values of $\lambda_{\text{SI}}$. The results are shown in \cref{tab:supp:HP-tuning}. As the upper bound, we choose $\lambda_{\text{SI}} = 1.0$ because the self-intersection loss should not be more important than the learning objective of the motion generation method. As the lower bound, we choose $\lambda_{\text{SI}} = 0.00001$ because we did not see an improvement beyond that value. Additionally, we investigate different loss formulations: An L2 loss, as in the main part of the paper, and an L1 loss.
\begin{table}[ht]
    \centering
    \resizebox{1.0\linewidth}{!}{
    \begin{tabular}{lcccccccc}
    \toprule
    \multirow{2}{2.cm}{\centering Method} &
    \multirow{2}{2.cm}{\centering R Precision top 1$\uparrow$} & 
    \multirow{2}{2.cm}{\centering R Precision top 2$\uparrow$} & 
    \multirow{2}{2.cm}{\centering R Precision top 3$\uparrow$} & 
    \multirow{2}{1.5cm}{\centering FID$\downarrow$} &
    \multirow{2}{2.5cm}{\centering Multimodal Dist$\downarrow$} &
    \multirow{2}{2cm}{\centering Diversity$\rightarrow$} &
    \multirow{2}{2cm}{\centering Multimodality$\uparrow$} &
    \multirow{2}{2cm}{\centering SI$\downarrow$} \\
    \\
    
    \midrule
        L1, $\lambda_{\text{SI}} = 1.0$ & $0.030^{\pm 0.002}$ & $0.063^{\pm 0.003}$ & $0.097^{\pm 0.003}$ & $85.101^{\pm 0.039}$ & $9.052^{\pm 0.029}$ & $0.661^{\pm 0.014}$ & $0.000^{\pm 0.000}$ & $0^{\pm 0}$ \\
        L1, $\lambda_{\text{SI}} = 0.1$ & $0.035^{\pm 0.003}$ & $0.071^{\pm 0.003}$ & $0.107^{\pm 0.004}$ & $82.460^{\pm 0.042}$ & $8.857^{\pm 0.031}$ & $0.639^{\pm 0.013}$ & $0.000^{\pm 0.000}$ & $846^{\pm 0}$ \\
        L1, $\lambda_{\text{SI}} = 0.01$ & $0.348^{\pm 0.006}$ & $0.518^{\pm 0.007}$ & $0.628^{\pm 0.006}$ & $2.352^{\pm 0.074}$ & $4.108^{\pm 0.020}$ & $8.632^{\pm 0.076}$ & $\underline{2.870^{\pm 0.071}}$ & $\mathbf{292^{\pm 9}}$ \\
        L1, $\lambda_{\text{SI}} = 0.001$ & $0.415^{\pm 0.006}$ & $0.602^{\pm 0.006}$ & $0.711^{\pm 0.007}$ & $0.748^{\pm 0.060}$ & $3.517^{\pm 0.022}$ & $9.281^{\pm 0.079}$ & $2.800^{\pm 0.077}$ & $4511^{\pm 21}$ \\
        L1, $\lambda_{\text{SI}} = 0.0001$ & $0.418^{\pm 0.004}$ & $0.615^{\pm 0.006}$ & $0.723^{\pm 0.007}$ & $0.606^{\pm 0.064}$ & $3.524^{\pm 0.021}$ & $\underline{9.578^{\pm 0.089}}$ & $2.752^{\pm 0.069}$ & $577^{\pm 20}$ \\
        L1, $\lambda_{\text{SI}} = 0.00001$ & $0.404^{\pm 0.006}$ & $0.595^{\pm 0.006}$ & $0.701^{\pm 0.004}$ & $1.017^{\pm 0.079}$ & $3.665^{\pm 0.023}$ & $9.264^{\pm 0.090}$ & $2.787^{\pm 0.080}$ & $608^{\pm 13}$ \\
    \midrule
        L2, $\lambda_{\text{SI}} = 1.0$ & $0.034^{\pm 0.003}$ & $0.067^{\pm 0.003}$ & $0.100^{\pm 0.003}$ & $79.416^{\pm 0.038}$ & $8.740^{\pm 0.029}$ & $0.590^{\pm 0.012}$ & $0.000^{\pm 0.000}$ & $1613^{\pm 0}$ \\
        L2, $\lambda_{\text{SI}} = 0.1$ & $0.313^{\pm 0.005}$ & $0.475^{\pm 0.006}$ & $0.583^{\pm 0.007}$ & $3.848^{\pm 0.101}$ & $4.409^{\pm 0.019}$ & $7.787^{\pm 0.070}$ & $2.790^{\pm 0.056}$ & $3088^{\pm 15}$ \\
        L2, $\lambda_{\text{SI}} = 0.01$ & $0.435^{\pm 0.005}$ & $0.628^{\pm 0.006}$ & $0.731^{\pm 0.006}$ & $\mathbf{0.265^{\pm 0.024}}$ & $3.462^{\pm 0.026}$ & $\mathbf{9.568^{\pm 0.086}}$ & $\mathbf{2.893^{\pm 0.075}}$ & $\underline{382^{\pm 9}}$ \\
        L2, $\lambda_{\text{SI}} = 0.001$ & $\mathbf{0.448^{\pm 0.008}}$ & $\mathbf{0.637^{\pm 0.007}}$ & $\mathbf{0.740^{\pm 0.007}}$ & $0.375^{\pm 0.025}$ & $\mathbf{3.351^{\pm 0.023}}$ & $9.634^{\pm 0.073}$ & $2.661^{\pm 0.062}$ & $432^{\pm 13}$ \\
        L2, $\lambda_{\text{SI}} = 0.0001$ & $0.430^{\pm 0.006}$ & $0.625^{\pm 0.006}$ & $0.728^{\pm 0.007}$ & $\mathbf{0.265^{\pm 0.024}}$ & $3.454^{\pm 0.026}$ & $9.659^{\pm 0.064}$ & $2.775^{\pm 0.068}$ & $452^{\pm 10}$ \\
        L2, $\lambda_{\text{SI}} = 0.00001$ & $\underline{0.440^{\pm 0.006}}$ & $\underline{0.635^{\pm 0.005}}$ & $\underline{0.736^{\pm 0.006}}$ & $\underline{0.287^{\pm 0.031}}$ & $\underline{3.361^{\pm 0.020}}$ & $9.607^{\pm 0.069}$ & $2.800^{\pm 0.068}$ & $490^{\pm 12}$ \\
    \midrule
        MDM~\cite{TRG+23-MDM} & $0.418^{\pm.005}$ & $0.604^{\pm .005}$ & $0.707^{\pm .004}$ & $0.489^{\pm .025}$ & $3.631^{\pm .023}$ & $9.449^{\pm .066}$ & $2.973^{\pm .111}$ & $619^{\pm 11}$ \\
        %Real & $0.511^{\pm .003}$ & $0.703^{\pm .003}$ & $0.797^{\pm .002}$ & $0.002^{\pm .000}$ & $2.974^{\pm .008}$ & $9.503^{\pm .065}$ & - & $446.912^{\pm 0.000}$ \\
    \bottomrule

    \end{tabular}
        }
    \caption{Hyperparameter tuning for $\lambda_{\text{SI}}$ on the HumanML3D~\cite{GZZ+22-humanml3d} dataset for SIA-MDM. All experiments are repeated for 20 random seeds. $\pm$ indicates the 95\% confidence interval. \textbf{Bold} indicates the best result, while \underline{underscore} indicates the second best. $\rightarrow$ indicates that closer to 'Real' ($9.503$) is better.}
    \label{tab:supp:HP-tuning}
\end{table}
\paragraph{SIA-MoMask}
We investigate the influence of the hyperparameter $\lambda_{\text{SI}}$ on the model performance of SIA-MoMask on HumanML3D~\cite{GZZ+22-humanml3d}. For this purpose, we train different versions of SIA-MoMask using varying values of $\lambda_{\text{SI}}$.
MoMask consists of a VQ-VAE, a masked transformer, and a residual transformer. The VQ-VAE learns a latent codebook, while the transformers are used to generate motions in that latent space. We integrate our novel self-intersection loss into the training of all three components of MoMask and investigate the following scenarios: 1) the self-intersection loss is integrated only into the training of the VQ-VAE (VQ-Only), 2) the self-intersection loss is only integrated into the training of the generative transformers while we use the pre-trained VQ-VAE of MoMask (Gen-Only), 3) the self-intersection loss is integrated into the training of all components (Both).
Additionally, we investigate the influence of the batch size during the training of the VQ-VAE. The publication of MoMask states that a batch size of $512$ is used during the training of the VQ-VAE. However, the provided checkpoint states that the VQ-VAE is trained with a batch size of $256$.
The results are shown in \cref{tab:supp:HP-tuning-MoMask}. As the upper bound, we choose $\lambda_{\text{SI}} = 0.1$ because higher values did not improve the model training. As the lower bound, we choose $\lambda_{\text{SI}} = 0.00001$ because we did not see an improvement beyond that value.
\begin{table*}[ht]
    \centering
    \resizebox{1.0\linewidth}{!}{
    \begin{tabular}{lcccccccc}
    \toprule
    \multirow{2}{2.cm}{\centering Method} &
    \multirow{2}{2.cm}{\centering R Precision top 1$\uparrow$} & 
    \multirow{2}{2.cm}{\centering R Precision top 2$\uparrow$} & 
    \multirow{2}{2.cm}{\centering R Precision top 3$\uparrow$} & 
    \multirow{2}{1.5cm}{\centering FID$\downarrow$} &
    \multirow{2}{2.5cm}{\centering Multimodal Dist$\downarrow$} &
    \multirow{2}{2cm}{\centering Diversity$\rightarrow$} &
    \multirow{2}{2cm}{\centering Multimodality$\uparrow$} &
    \multirow{2}{2cm}{\centering SI$\downarrow$} \\
    \\
    
    \midrule
        VQ-Only, $\lambda_{\text{SI}} = 0.1$ & $0.515^{\pm 0.003}$ & $0.709^{\pm 0.003}$ & $0.806^{\pm 0.003}$ & $0.094^{\pm 0.005}$ & $2.977^{\pm 0.010}$ & $9.621^{\pm 0.087}$ & $1.200^{\pm 0.040}$ & $309^{\pm 2}$ \\
        VQ-Only, $\lambda_{\text{SI}} = 0.01$ & $0.506^{\pm 0.002}$ & $0.699^{\pm 0.002}$ & $0.797^{\pm 0.002}$ & $0.106^{\pm 0.004}$ & $3.056^{\pm 0.008}$ & $9.643^{\pm 0.079}$ & $1.280^{\pm 0.041}$ & $312^{\pm 3}$ \\
        VQ-Only, $\lambda_{\text{SI}} = 0.001$ & $\mathbf{0.525^{\pm 0.003}}$ & $\mathbf{0.718^{\pm 0.002}}$ & \underline{$0.812^{\pm 0.002}$} & $0.072^{\pm 0.003}$ & $2.957^{\pm 0.007}$ & $9.729^{\pm 0.098}$ & $1.218^{\pm 0.035}$ & $723^{\pm 2}$ \\
        VQ-Only, $\lambda_{\text{SI}} = 0.0001$ & $0.502^{\pm 0.002}$ & $0.693^{\pm 0.002}$ & $0.793^{\pm 0.002}$ & $0.082^{\pm 0.003}$ & $3.051^{\pm 0.007}$ & $9.593^{\pm 0.080}$ & $1.288^{\pm 0.036}$ & $298^{\pm 2}$ \\
        VQ-Only, $\lambda_{\text{SI}} = 0.00001$ & $0.511^{\pm 0.003}$ & $0.705^{\pm 0.002}$ & $0.800^{\pm 0.002}$ & $0.076^{\pm 0.003}$ & $3.035^{\pm 0.008}$ & $9.569^{\pm 0.080}$ & $1.272^{\pm 0.039}$ & $341^{\pm 2}$ \\
        VQ-Only, $\lambda_{\text{SI}} = 0.000001$ & $0.501^{\pm 0.003}$ & $0.699^{\pm 0.003}$ & $0.798^{\pm 0.002}$ & $0.085^{\pm 0.005}$ & $3.070^{\pm 0.007}$ & \underline{$9.473^{\pm 0.084}$} & \underline{$1.337^{\pm 0.037}$} & $352^{\pm 3}$ \\
    \midrule        
        VQ-Only (256), $\lambda_{\text{SI}} = 0.1$ & $0.514^{\pm 0.002}$ & $0.707^{\pm 0.002}$ & $0.800^{\pm 0.002}$ & $0.077^{\pm 0.003}$ & $3.009^{\pm 0.007}$ & $9.651^{\pm 0.089}$ & $1.212^{\pm 0.042}$ & $1556^{\pm 2}$ \\
        VQ-Only (256), $\lambda_{\text{SI}} = 0.01$ & $\mathbf{0.525^{\pm 0.003}}$ & \underline{$0.717^{\pm 0.003}$} & $\mathbf{0.813^{\pm 0.002}}$ & $0.068^{\pm 0.002}$ & $\mathbf{2.933^{\pm 0.006}}$ & $9.691^{\pm 0.092}$ & $1.198^{\pm 0.041}$ & \underline{$290^{\pm 2}$} \\
        VQ-Only (256), $\lambda_{\text{SI}} = 0.001$ & $0.497^{\pm 0.002}$ & $0.693^{\pm 0.002}$ & $0.794^{\pm 0.002}$ & $0.147^{\pm 0.005}$ & $3.081^{\pm 0.009}$ & $9.466^{\pm 0.084}$ & $\mathbf{1.377^{\pm 0.041}}$ & $\mathbf{287^{\pm 3}}$ \\
        VQ-Only (256), $\lambda_{\text{SI}} = 0.0001$ & $0.522^{\pm 0.002}$ & $0.716^{\pm 0.003}$ & $0.810^{\pm 0.002}$ & $0.068^{\pm 0.004}$ & $2.950^{\pm 0.008}$ & $9.710^{\pm 0.082}$ & $1.179^{\pm 0.037}$ & $394^{\pm 2}$ \\
        VQ-Only (256), $\lambda_{\text{SI}} = 0.00001$ & $0.517^{\pm 0.002}$ & $0.711^{\pm 0.002}$ & $0.808^{\pm 0.002}$ & $0.082^{\pm 0.003}$ & $2.991^{\pm 0.007}$ & $9.702^{\pm 0.082}$ & $1.252^{\pm 0.045}$ & $333^{\pm 2}$ \\
        VQ-Only (256), $\lambda_{\text{SI}} = 0.000001$ & $0.520^{\pm 0.003}$ & $0.709^{\pm 0.003}$ & $0.805^{\pm 0.003}$ & $0.063^{\pm 0.002}$ & $2.973^{\pm 0.007}$ & $9.668^{\pm 0.080}$ & $1.179^{\pm 0.043}$ & $364^{\pm 2}$ \\
    \midrule
        Gen-Only, $\lambda_{\text{SI}} = 0.1$ & $0.515^{\pm 0.002}$ & $0.709^{\pm 0.003}$ & $0.806^{\pm 0.002}$ & $0.077^{\pm 0.003}$ & $2.988^{\pm 0.007}$ & $9.677^{\pm 0.079}$ & $1.269^{\pm 0.040}$ & $300^{\pm 2}$ \\
        Gen-Only, $\lambda_{\text{SI}} = 0.01$ & $0.519^{\pm 0.002}$ & $0.711^{\pm 0.002}$ & $0.807^{\pm 0.002}$ & $0.063^{\pm 0.003}$ & $2.972^{\pm 0.008}$ & $9.702^{\pm 0.076}$ & $1.263^{\pm 0.039}$ & $291^{\pm 2}$ \\
        Gen-Only, $\lambda_{\text{SI}} = 0.001$ & $0.518^{\pm 0.002}$ & $0.710^{\pm 0.003}$ & $0.807^{\pm 0.002}$ & $0.070^{\pm 0.002}$ & $2.981^{\pm 0.008}$ & $9.677^{\pm 0.083}$ & $1.271^{\pm 0.038}$ & $297^{\pm 2}$ \\
        Gen-Only, $\lambda_{\text{SI}} = 0.0001$ & $0.521^{\pm 0.002}$ & $0.713^{\pm 0.002}$ & $0.809^{\pm 0.002}$ & $0.063^{\pm 0.003}$ & $2.962^{\pm 0.009}$ & $9.731^{\pm 0.086}$ & $1.247^{\pm 0.041}$ & $302^{\pm 2}$ \\
        Gen-Only, $\lambda_{\text{SI}} = 0.00001$ & $0.521^{\pm 0.003}$ & $0.713^{\pm 0.002}$ & $0.808^{\pm 0.002}$ & $0.074^{\pm 0.004}$ & $2.968^{\pm 0.006}$ & $9.672^{\pm 0.107}$ & $1.206^{\pm 0.043}$ & $313^{\pm 2}$ \\
        Gen-Only, $\lambda_{\text{SI}} = 0.000001$ & $0.519^{\pm 0.003}$ & $0.714^{\pm 0.002}$ & $0.809^{\pm 0.002}$ & $0.074^{\pm 0.004}$ & $2.967^{\pm 0.006}$ & $9.676^{\pm 0.099}$ & $1.207^{\pm 0.040}$ & $314^{\pm 3}$ \\
    \midrule 
        Both, $\lambda_{\text{SI}} = 0.1$ & $0.506^{\pm 0.003}$ & $0.701^{\pm 0.002}$ & $0.800^{\pm 0.002}$ & $0.143^{\pm 0.005}$ & $3.035^{\pm 0.009}$ & $9.542^{\pm 0.096}$ & $1.272^{\pm 0.038}$ & $1559^{\pm 3}$ \\
        Both, $\lambda_{\text{SI}} = 0.01$ & $0.521^{\pm 0.003}$ & $0.716^{\pm 0.002}$ & \underline{$0.812^{\pm 0.002}$} & $0.079^{\pm 0.003}$ & \underline{$2.947^{\pm 0.009}$} & $9.733^{\pm 0.097}$ & $1.187^{\pm 0.038}$ & $302^{\pm 2}$ \\
        Both, $\lambda_{\text{SI}} = 0.001$ & $0.506^{\pm 0.003}$ & $0.700^{\pm 0.003}$ & $0.797^{\pm 0.002}$ & $0.185^{\pm 0.007}$ & $3.048^{\pm 0.007}$ & $\mathbf{9.498^{\pm 0.080}}$ & $1.300^{\pm 0.041}$ & $354^{\pm 3}$ \\
        Both, $\lambda_{\text{SI}} = 0.0001$ & $0.522^{\pm 0.003}$ & $0.715^{\pm 0.002}$ & \underline{$0.812^{\pm 0.002}$} & $0.082^{\pm 0.004}$ & $2.949^{\pm 0.008}$ & $9.747^{\pm 0.092}$ & $1.160^{\pm 0.039}$ & $397^{\pm 3}$ \\
        Both, $\lambda_{\text{SI}} = 0.00001$ & $0.512^{\pm 0.002}$ & $0.704^{\pm 0.002}$ & $0.800^{\pm 0.002}$ & $0.117^{\pm 0.005}$ & $3.024^{\pm 0.007}$ & $9.701^{\pm 0.099}$ & $1.279^{\pm 0.043}$ & $373^{\pm 3}$ \\
        Both, $\lambda_{\text{SI}} = 0.000001$ & $0.521^{\pm 0.003}$ & $0.715^{\pm 0.002}$ & $0.810^{\pm 0.002}$ & $0.084^{\pm 0.003}$ & $2.970^{\pm 0.009}$ & $9.665^{\pm 0.089}$ & $1.229^{\pm 0.033}$ & $400^{\pm 2}$ \\
    \midrule
        Both (256), $\lambda_{\text{SI}} = 0.1$ & $0.515^{\pm 0.002}$ & $0.713^{\pm 0.003}$ & $0.810^{\pm 0.002}$ & $0.092^{\pm 0.004}$ & $2.965^{\pm 0.008}$ & $9.609^{\pm 0.090}$ & $1.218^{\pm 0.036}$ & $333^{\pm 2}$ \\
        Both (256), $\lambda_{\text{SI}} = 0.01$ & $0.508^{\pm 0.002}$ & $0.702^{\pm 0.002}$ & $0.801^{\pm 0.002}$ & $0.127^{\pm 0.004}$ & $3.040^{\pm 0.006}$ & $9.593^{\pm 0.079}$ & $1.315^{\pm 0.037}$ & $299^{\pm 2}$ \\
        Both (256), $\lambda_{\text{SI}} = 0.001$ & \underline{$0.524^{\pm 0.002}$} & $\mathbf{0.718^{\pm 0.002}}$ & $\mathbf{0.813^{\pm 0.002}}$ & $0.074^{\pm 0.003}$ & \underline{$2.947^{\pm 0.008}$} & $9.777^{\pm 0.079}$ & $1.176^{\pm 0.038}$ & $734^{\pm 3}$ \\
        Both (256), $\lambda_{\text{SI}} = 0.0001$ & $0.514^{\pm 0.002}$ & $0.706^{\pm 0.003}$ & $0.803^{\pm 0.002}$ & \underline{$0.060^{\pm 0.003}$} & $3.018^{\pm 0.008}$ & $9.674^{\pm 0.087}$ & $1.279^{\pm 0.041}$ & $363^{\pm 3}$ \\
        Both (256), $\lambda_{\text{SI}} = 0.00001$ & $0.514^{\pm 0.002}$ & $0.707^{\pm 0.002}$ & $0.803^{\pm 0.002}$ & $0.064^{\pm 0.002}$ & $3.000^{\pm 0.008}$ & $9.588^{\pm 0.088}$ & $1.244^{\pm 0.044}$ & $372^{\pm 3}$ \\
        Both (256), $\lambda_{\text{SI}} = 0.000001$ & $0.513^{\pm 0.002}$ & $0.707^{\pm 0.002}$ & $0.804^{\pm 0.002}$ & $0.082^{\pm 0.004}$ & $3.013^{\pm 0.008}$ & $9.614^{\pm 0.096}$ & $1.257^{\pm 0.041}$ & $348^{\pm 2}$ \\
    \midrule
        MoMask~\cite{GMJ+24-MoMask} & $0.521^{\pm 0.002}$ & $0.713^{\pm 0.002}$ & $0.807^{\pm 0.002}$ & $\mathbf{0.045^{\pm 0.002}}$ & $2.958^{\pm 0.008}$ & $9.624^{\pm 0.080}$ & $1.241^{\pm 0.040}$ & $316^{\pm 2}$ \\
    \bottomrule

    \end{tabular}
        }
    \caption{Hyperparameter tuning for $\lambda_{\text{SI}}$ on the HumanML3D~\cite{GZZ+22-humanml3d} dataset for SIA-MoMask. All experiments are repeated for 20 random seeds. $\pm$ indicates the 95\% confidence interval. \textbf{Bold} indicates the best result, while \underline{underscore} indicates the second best. $\rightarrow$ indicates that closer to 'Real' ($9.503$) is better.}
    \label{tab:supp:HP-tuning-MoMask}
\end{table*}

\section{Recovering SMPL from HumanML3D}
\label{sec:supp:hml-to-smpl}

\begin{figure}[ht]
\centering
\subfigure[HumanML3D]{
    \includegraphics[width=0.4\textwidth]{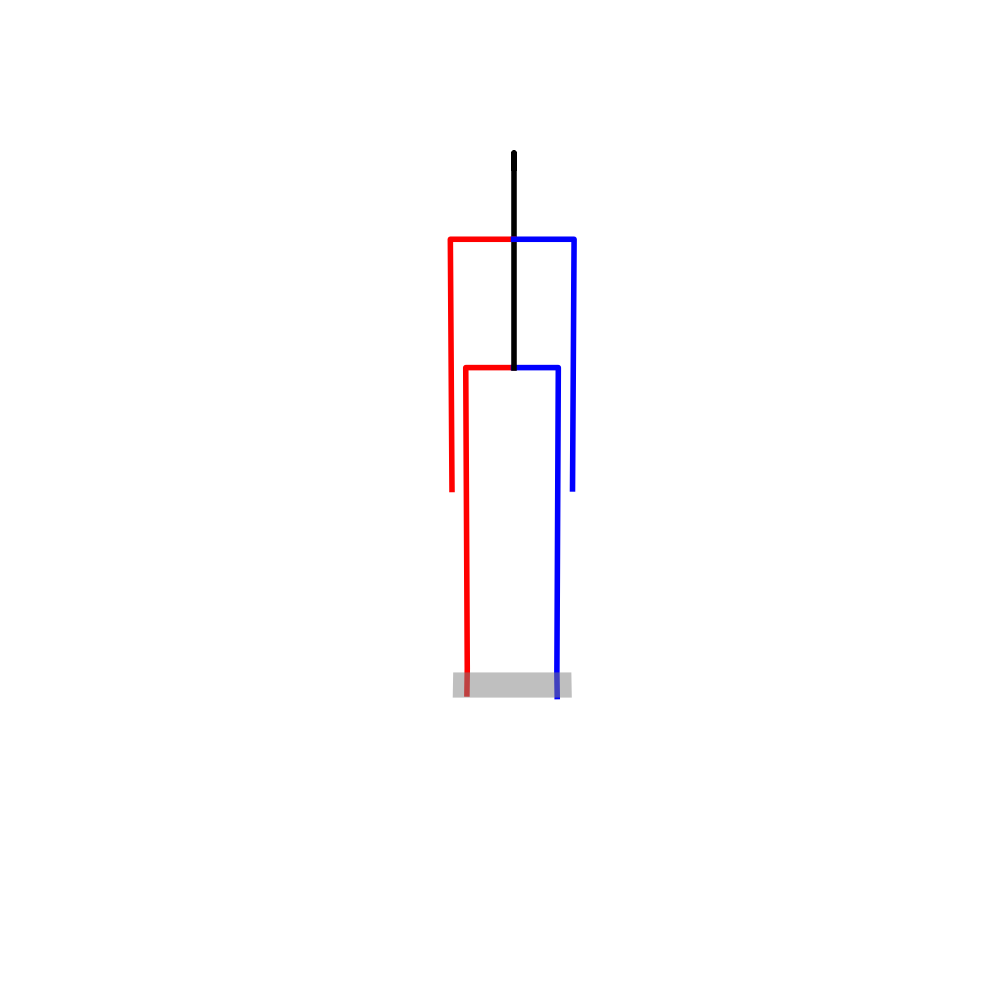}
    \label{fig:supp:HumanML3D-skeleton}
}
\subfigure[SMPL]{
    \includegraphics[width=0.4\textwidth]{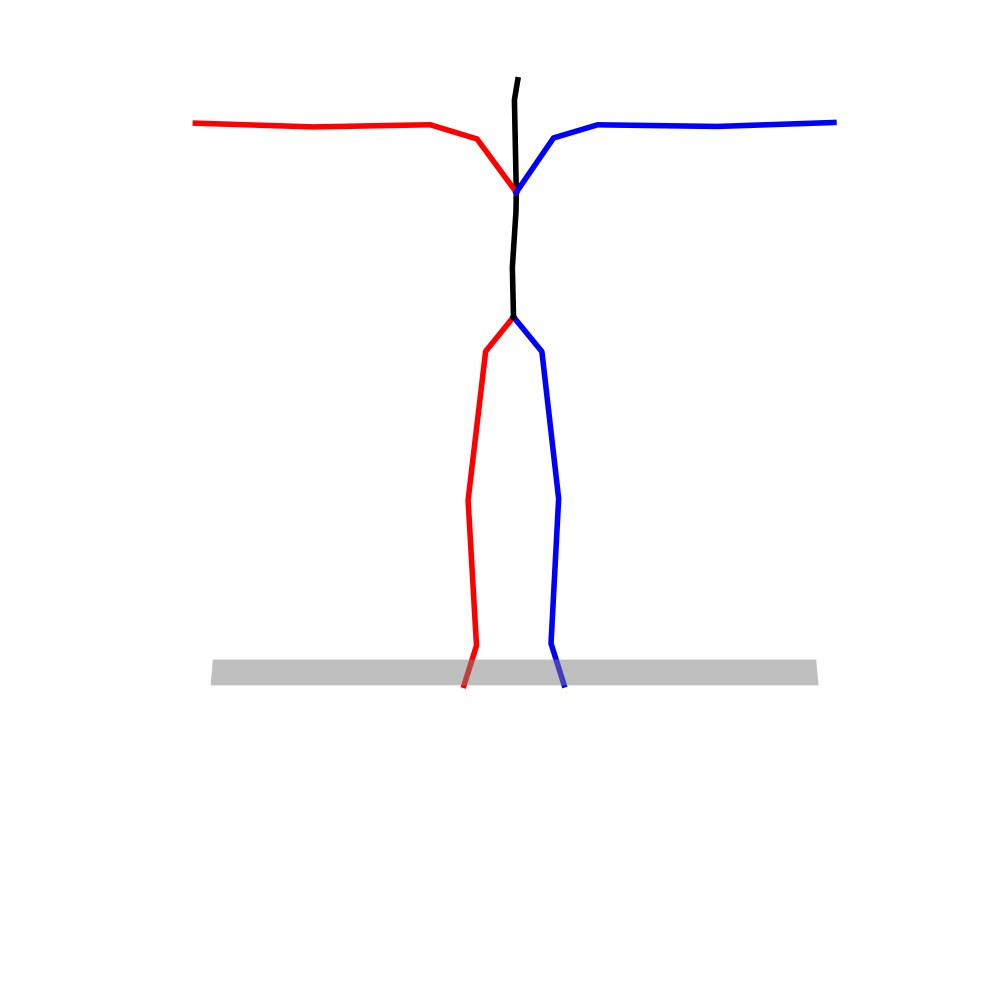}
    \label{fig:supp:SMPL-skeleton}
}      
\caption{Skeletons in the default pose of HumanML3D and SMPL.}
\label{fig:supp:default-skeletons}
\end{figure}
We base our sphere proxy on the SMPL model and use the SMPL skeleton structure to define the poses of our sphere proxy. However, most recent text-to-motion generation methods use the motion representation proposed by HumanML3D~\cite{GZZ+22-humanml3d}. We recover the SMPL joint rotations from that motion representation to apply our sphere proxy for the text-to-motion task. A common approach is to apply SMPLify~\cite{BKL+16-SMPLify} to the generated motions to obtain the SMPL shape parameters and joint rotations for visualization purposes. However, SMPLify takes several minutes to process a single motion, inhibiting its use during the training of motion generation methods. Instead, we split the motion recovery process into two parts.

The first step is to recover the SMPL shape parameters. HumanML3D is built upon the AMASS~\cite{MGT+19-AMASS} dataset, which contains motions by many people with different shape parameters. HumanML3D converts these motions to 3D keypoints and retargets them to a target skeleton. This target skeleton is simply the skeleton of one of the motions from AMASS~\cite{MGT+19-AMASS}. The target motion is represented as a male SMPL model, and the shape parameters could be obtained from the ground truth data. Instead, since the motions in HumanML3D can be seen as genderless, we use SMPLify on the target motion, fitting a genderless SMPL-H model to it. SMPLify is frame-based, resulting in different shape parameters for every motion frame. We average the shape parameters over all frames and obtain the shape parameters for the HumanML3D standard skeleton.

The next step is to recover the SMPL joint rotations from the HumanML3D motion representation, which has the following form:
\begin{equation}
    \vx^i = (\dot{r}^a, \dot{r}^x, \dot{r}^z, r^y, \vj^p, \vj^r, \vj^v, \vc^f).
\end{equation}
Here, $\dot{r}^a \in \mathbb{R}$ is the root angular velocity along the y-axis, $\dot{r}^x \in \mathbb{R}$ is the root linear velocity in x-direction, $\dot{r}^z \in \mathbb{R}$ is the root linear velocity in z-direction, $r^y \in \mathbb{R}$ is the root height, $\vj^p \in \mathbb{R}^{3J}$, $\vj^r \in \mathbb{R}^{6J}$, and $\vj^v \in \mathbb{R}^{3J}$ are the local joint positions, rotations, and velocities with $J$ denoting the number of joints, and $\vc^f \in \mathbb{R}^{4}$ are foot ground contact features. Joint rotations are represented in the 6D continuous representation of Zhou~et~al.~\cite{ZBL+19-rotrep}. The motions in HumanML3D follow the skeleton structure of SMPL-H~\cite{LMR+15-SMPL,RTB17-MANO}, disregarding the hand joints, resulting in a skeleton structure with $J=22$ joints.
While the HumanML3D pose representation contains joint rotations, they are inconsistent with SMPL. First, the skeleton structure differs for both approaches, as shown in \cref{fig:supp:default-skeletons}. Because of this difference, the same rotation angle value results in different joint locations. However, accounting for the different skeletons is insufficient, as both representations define the joint angles differently. For HumanML3D, the rotation of a joint is calculated by first calculating the vector from that joint to its \textbf{parent} joint according to the kinematic tree. Next, the corresponding vector is obtained from the default skeleton (all joint rotations equal zero) as shown in \cref{fig:supp:HumanML3D-skeleton}. The joint rotation is then defined as the angle between both vectors.
In contrast, SMPL uses the vector of the joint to its \textbf{child} to define the joint angles. As an example, consider your elbow. HumanML3D defines the elbow rotation by the vector from the shoulder to the elbow. This way, the elbow rotation has a direct effect on the location of the elbow itself. SMPL defines the elbow rotation by the vector from the elbow to the wrist. This way, the elbow rotation only affects the position of subsequent joints but not the position of the elbow itself.
Keeping this discussion in mind, we obtain the SMPL parameters as follows. First, the global joint locations are obtained from the HumanML3D representation using the local joint positions, the root height, and the root angular and linear velocity. Next, the vectors connecting adjacent joints according to the kinematic tree are calculated. Finally, the joint rotations are computed as the angle between the current and default vectors, as shown in \cref{fig:supp:SMPL-skeleton}.
This procedure is fast and differentiable, which allows us to recover SMPL joint rotations from the HumanML3D pose representation during the training of motion generation methods and backpropagate our self-intersection loss through the joint rotation recovery.
Furthermore, this approach decouples the generation of motion from the generation of the identity of the moving person, which makes it possible to vary the SMPL shape parameter to let different persons execute the generated motion. Using different SMPL shape parameters results in a discrepancy between the generated 3D joint locations and the joint locations of the SMPL model, however, the motion (i.e. how the individual body parts move) stays the same.

\section{Ablation Results}
\label{sec:supp:abblation}

In \cref{subsec:ex:ablations}, we compare the memory usage and runtime of the sphere proxy and the mesh-based loss for different numbers of frames to apply the loss to. For completeness, \cref{tab:supp:abl_collision_method} reports the metrics calculated for the resulting models.
\textit{SI} improves regardless of loss type and number of frames in almost all cases, highlighting the benefit of adding a self-intersection loss to the model training.
A drop in performance can be observed at 4 frames for both methods, which is most likely due to the sampling of the frames to which the loss is applied. We do not investigate different sampling strategies as our sphere proxy enables us to apply the loss to all frames.

\begin{table*}[ht]
    \centering
    \resizebox{1.0\linewidth}{!}{
    \begin{tabular}{lcccccccc}
    \toprule
    \multirow{2}{2.cm}{\centering Method} &
    \multirow{2}{2.cm}{\centering R Precision top 1$\uparrow$} & 
    \multirow{2}{2.cm}{\centering R Precision top 2$\uparrow$} & 
    \multirow{2}{2.cm}{\centering R Precision top 3$\uparrow$} & 
    \multirow{2}{1.5cm}{\centering FID$\downarrow$} &
    \multirow{2}{2.5cm}{\centering Multimodal Dist$\downarrow$} &
    \multirow{2}{2cm}{\centering Diversity$\rightarrow$} &
    \multirow{2}{2cm}{\centering MultiModality$\uparrow$} &
    \multirow{2}{2cm}{\centering SI$\downarrow$} \\
    \\
    
    \midrule
        Spheres (1 frames) & $0.429^{\pm 0.005}$ & $0.616^{\pm 0.005}$ & $0.717^{\pm 0.007}$ & $0.546^{\pm 0.039}$ & $3.532^{\pm 0.025}$ & $9.676^{\pm 0.071}$ & $2.766^{\pm 0.059}$ & $559^{\pm 10}$ \\
        Spheres (2 frames) & $0.432^{\pm 0.006}$ & $0.632^{\pm 0.008}$ & $0.736^{\pm 0.007}$ & $0.527^{\pm 0.043}$ & $3.447^{\pm 0.020}$ & $9.361^{\pm 0.080}$ & $2.742^{\pm 0.061}$ & $501^{\pm 19}$ \\
        Spheres (4 frames) & $0.418^{\pm 0.005}$ & $0.612^{\pm 0.007}$ & $0.721^{\pm 0.007}$ & $0.595^{\pm 0.030}$ & $3.520^{\pm 0.028}$ & $9.297^{\pm 0.077}$ & $2.619^{\pm 0.066}$ & $327^{\pm 14}$ \\
        Spheres (8 frames) & $0.435^{\pm 0.005}$ & $0.631^{\pm 0.005}$ & $0.735^{\pm 0.005}$ & $0.480^{\pm 0.022}$ & $3.432^{\pm 0.022}$ & $9.430^{\pm 0.090}$ & $2.591^{\pm 0.070}$ & $345^{\pm 13}$ \\    
        Spheres (16 frames) & $0.409^{\pm 0.006}$ & $0.596^{\pm 0.007}$ & $0.702^{\pm 0.005}$ & $0.536^{\pm 0.051}$ & $3.667^{\pm 0.027}$ & $9.405^{\pm 0.051}$ & $2.829^{\pm 0.072}$ & $402^{\pm 11}$ \\
        Spheres (196 frames) & $0.435^{\pm 0.005}$ & $0.628^{\pm 0.006}$ & $0.731^{\pm 0.006}$ & $0.265^{\pm 0.024}$ & $3.462^{\pm 0.026}$ & $9.568^{\pm 0.086}$ & $2.893^{\pm 0.075}$ & $382^{\pm 09}$ \\
    \midrule
        Mesh (1 frames) & $0.422^{\pm 0.006}$ & $0.618^{\pm 0.007}$ & $0.725^{\pm 0.005}$ & $0.481^{\pm 0.035}$ & $3.543^{\pm 0.025}$ & $9.590^{\pm 0.070}$ & $2.861^{\pm 0.071}$ & $416^{\pm 13}$ \\
        Mesh (2 frames) & $0.422^{\pm 0.007}$ & $0.613^{\pm 0.008}$ & $0.720^{\pm 0.007}$ & $0.480^{\pm 0.042}$ & $3.558^{\pm 0.031}$ & $9.545^{\pm 0.083}$ & $2.789^{\pm 0.078}$ & $504^{\pm 20}$ \\
        Mesh (4 frames) & $0.412^{\pm 0.007}$ & $0.602^{\pm 0.007}$ & $0.708^{\pm 0.006}$ & $0.741^{\pm 0.060}$ & $3.615^{\pm 0.020}$ & $9.310^{\pm 0.067}$ & $2.694^{\pm 0.072}$ & $751^{\pm 25}$ \\
        Mesh (8 frames) & $0.421^{\pm 0.007}$ & $0.615^{\pm 0.007}$ & $0.723^{\pm 0.007}$ & $0.418^{\pm 0.035}$ & $3.547^{\pm 0.027}$ & $9.348^{\pm 0.080}$ & $2.699^{\pm 0.079}$ & $609^{\pm 17}$ \\
        Mesh (16 frames) & $0.411^{\pm 0.006}$ & $0.607^{\pm 0.006}$ & $0.715^{\pm 0.004}$ & $0.498^{\pm 0.046}$ & $3.545^{\pm 0.019}$ & $9.600^{\pm 0.076}$ & $2.695^{\pm 0.071}$ & $592^{\pm 12}$ \\
        Mesh (196 frames) & - & - & - & - & - & - & - & - \\
    \midrule
        MDM~\cite{TRG+23-MDM} & $0.418^{\pm.005}$ & $0.604^{\pm 0.005}$ & $0.707^{\pm 0.004}$ & $0.489^{\pm 0.025}$ & $3.631^{\pm 0.023}$ & $9.449^{\pm 0.066}$ & $2.973^{\pm 0.111}$ & $619^{\pm 11}$ \\
    \bottomrule

    \end{tabular}
    }
    \caption{Comparison of the self-intersection loss calculated using meshes and the sphere proxy on HumanML3D~\cite{GZZ+22-humanml3d}. All experiments are repeated for 20 random seeds. $\pm$ indicates the 95\% confidence interval. $\rightarrow$ indicates that closer to 'Real' is better}
    \label{tab:supp:abl_collision_method}
\end{table*}

\section{Sphere Proxy Evaluation}
\label{sec:supp:sphere_proxy}

\paragraph{Hyperparameter tuning}
\begin{table}[ht]
\begin{minipage}[b]{0.49\linewidth}
\centering
\includegraphics[width=0.98\textwidth]{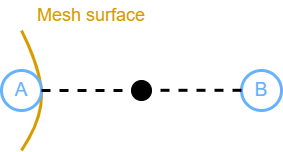}
  \captionof{figure}{Motivation for the emptiness loss. The curved line represents the surface of the mesh, which should be approximated by a set of spheres. The filled dot in the middle represents a sampled point with some SDF value for the given mesh. Spheres A and B have the same distance to the sampled point, resulting in valid placements of these spheres. However, as we want to approximate the surface of the mesh, only sphere A is desirable.}
  \label{fig:supp:SDF-ambiguity}
\end{minipage}\hfill
\begin{minipage}[b]{0.49\linewidth}
\centering
\includegraphics[width=0.98\textwidth]{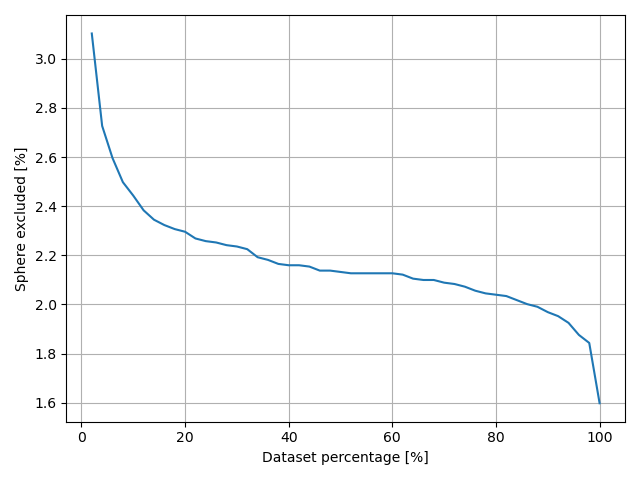}
  \captionof{figure}{Evaluation of collision reduction. Sphere intersections for all poses in the HumanML3D training set are recorded. The x-axis represents the percentage of poses for which sphere pairs intersect. The y-axis shows the percentage of spheres removed from the self-intersection loss calculation. We choose 90\%, corresponding to a 2\% reduction in spheres checked for intersections.}
  \label{fig:supp:sphere-reduction}
\end{minipage}
\end{table}

\begin{figure}[ht]
  \subfigure[Sphere proxy with $\lambda_{\text{emptiness}}=0.0$ and $\lambda_{\text{SI}}=0.0$.]{
    \includegraphics[width=0.4\textwidth]{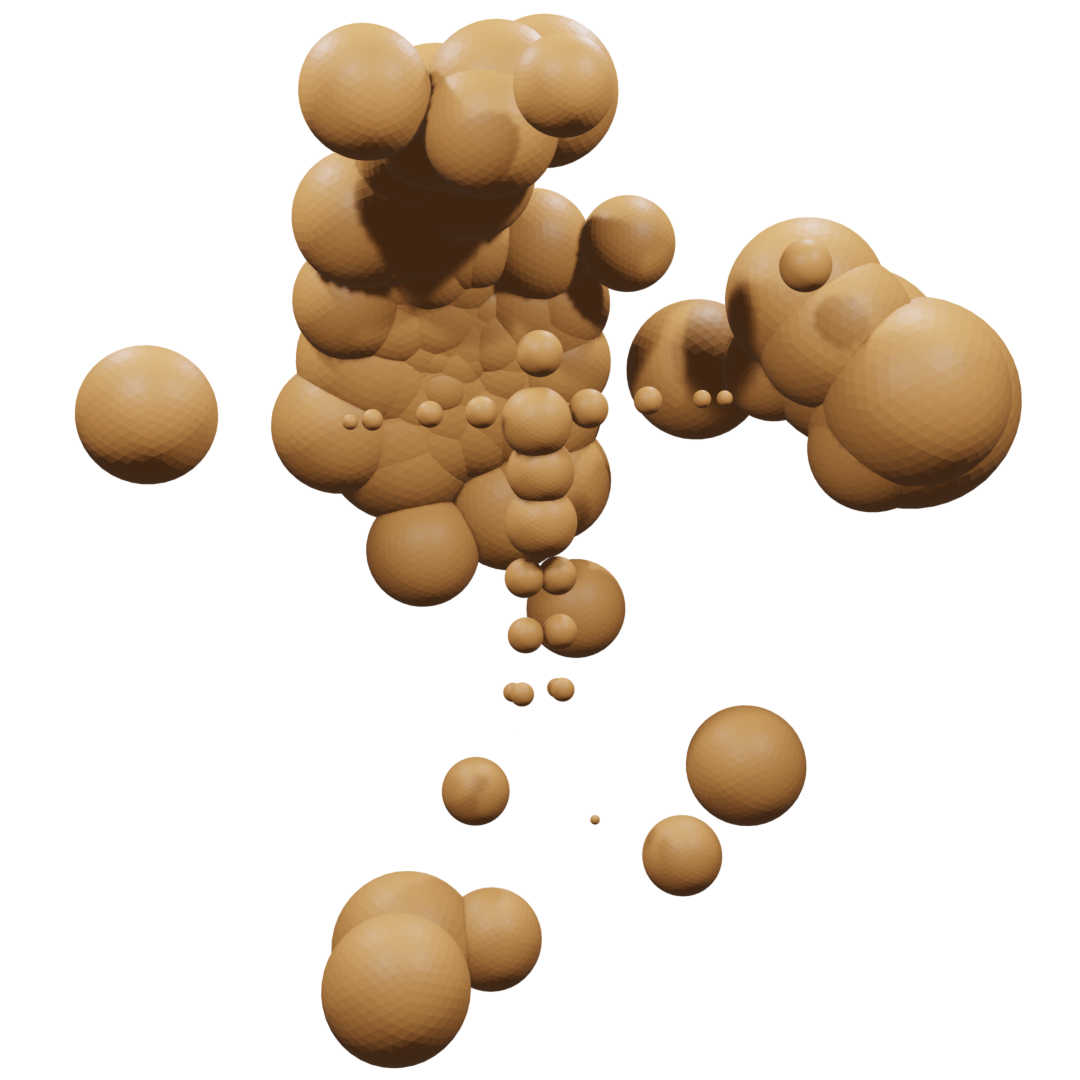}
  }
  \hspace{0.1\textwidth}
  \subfigure[Sphere proxy with $\lambda_{\text{emptiness}}=10.0$ and $\lambda_{\text{SI}}=0.1$.]{
    \includegraphics[width=0.4\textwidth]{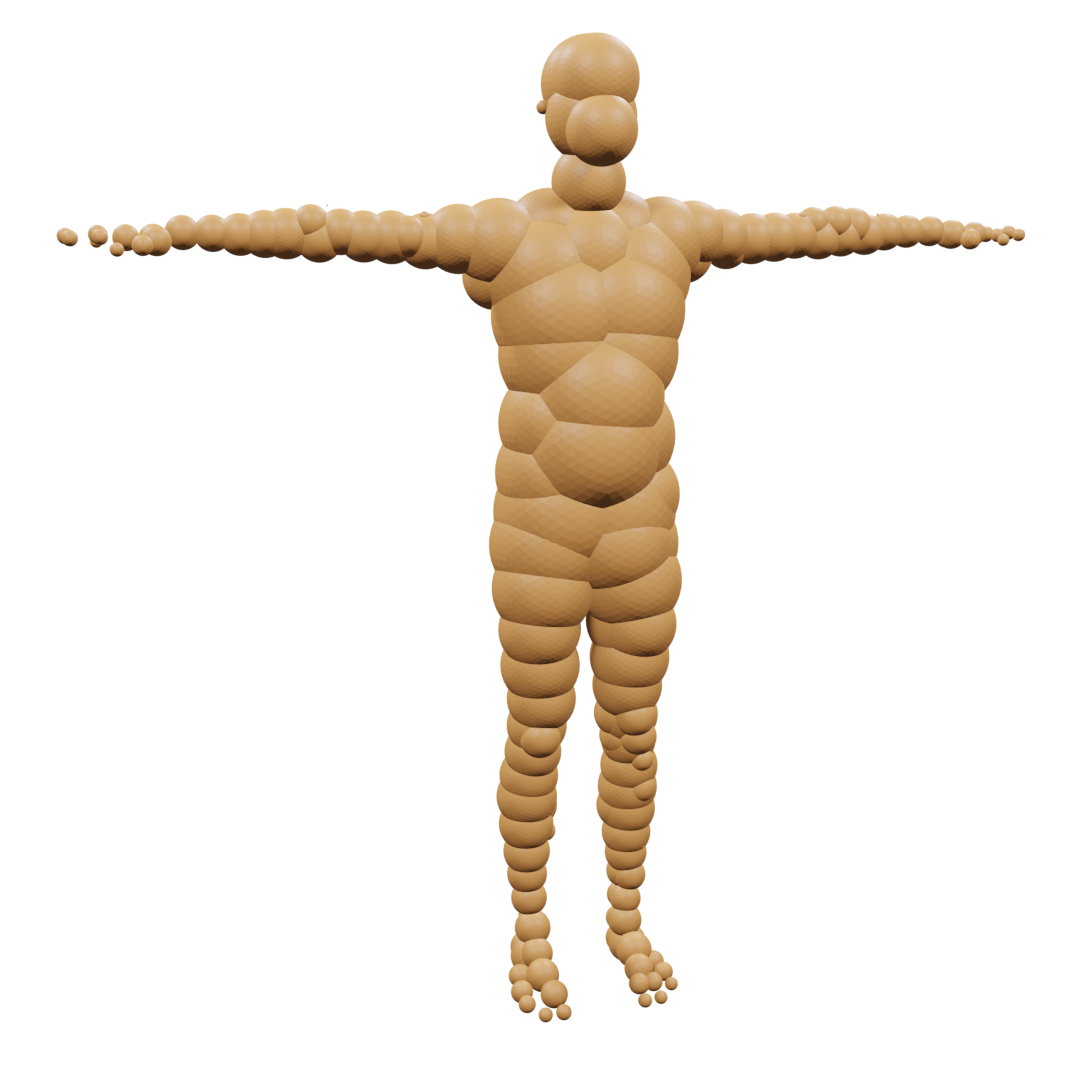}
  }
  \caption{Influence of the sphere proxy loss hyperparameters. For $\lambda_{\text{emptiness}}=0.0$ and $\lambda_{\text{SI}}=0.0$, the rough shape of the human can be recognized in the middle. However, most spheres are outside the boundaries of the original mesh, which makes the approximation invalid. For $\lambda_{\text{emptiness}}=10.0$ and $\lambda_{\text{SI}}=0.1$, a valid approximation of the human geometry is achieved.}
  \label{fig:supp:NoEmptiness}
\end{figure}
The loss of the sphere proxy contains two hyperparameters: $\lambda_{\text{emptiness}}$ and $\lambda_{\text{SI}}$. \cref{fig:supp:SDF-ambiguity} provides an additional motivation for the emptiness loss. We evaluate different settings of these hyperparameters in \cref{tab:supp:hyperparameters-sp}. We train the sphere proxies as described in \cref{subsec:ex:implementation}, and we also sample $150$ random SMPL~\cite{LMR+15-SMPL} shape parameters and compute the corresponding SMPL meshes, which serve as our test set. We use two metrics to compare the methods.
First, we evaluate how well the sphere proxy approximates the surface of the mesh and call this metric \textit{Surface}. We calculate the SDF values of the sphere proxy for all vertices of the mesh as in \cref{eq:sdf_sphere_proxy}. Then, we take the absolute value of the SDF values and take the sum over them. Formally, the metric is given by
\begin{equation}
    \text{Surface} = \sum_{v=1}^V |d_{S}^v|,
\end{equation}
where $V$ are the vertices of the mesh with $V=6890$ for the SMPL~\cite{LMR+15-SMPL} model. Dividing \textit{Surface} by $V$ results in the mean per-vertex deviation for the given sphere proxy in units of meters.
Second, we evaluate how well the sphere proxy approximates the volume of the mesh and call this metric \textit{VolDev}. We place the sphere proxy and the SMPL mesh at the origin and scale them to fit into a sphere with a radius of $1$~m. Similar to \cref{sec:experiments}, we approximate the volume of the sphere proxy and the mesh using voxels. We have a smaller test set than the human motion generation evaluation, so we can computationally afford to use smaller voxels with edge lengths $0.02$~cm. Let $\text{Vol}_{\text{SP}}$ and $\text{Vol}_{\text{Mesh}}$ be the approximated volumes of the sphere proxy and the mesh, respectively. With them, we define the metric \textit{VolDev} as the deviation of the approximated sphere proxy volume from the approximated mesh volume:
\begin{equation}
    \text{VolDev} = \left|\frac{\text{Vol}_{\text{Mesh}} - \text{Vol}_{\text{SP}}}{\text{Vol}_{\text{Mesh}}}\right|.
\end{equation}
We calculate both metrics for all SMPL shapes in the test set and report the mean and 95\% confidence interval.
The results indicate that the emptiness loss is essential to train the sphere proxy. Without it, the surface and the volume are not properly approximated. In addition, small values for $\lambda_{\text{IS}}$ result in a small improvement compared to not using the loss. An example sphere proxy with $\lambda_{\text{emptiness}}=0.0$ and $\lambda_{\text{SI}}=0.0$ is shown in \cref{fig:supp:NoEmptiness}. Most spheres are outside the boundaries of the original mesh, making the approximation invalid.

\begin{table}[ht]
    \centering
    \resizebox{1.0\linewidth}{!}{
    \begin{tabular}{ccc}
    \toprule
    \multirow{2}{2.cm}{\centering Method} &
    \multirow{2}{2.cm}{\centering Surface $\downarrow$} & 
    \multirow{2}{2.cm}{\centering VolDev $\downarrow$} \\
    \\
    
    \midrule
        $\lambda_{\text{emptiness}} = 0.0$, $\lambda_{\text{SI}} = 0.0$ & $1870^{\pm 12}$ & $1.597^{\pm 0.191}$ \\ 
        $\lambda_{\text{emptiness}} = 5.0$, $\lambda_{\text{SI}} = 0.0$ & $39^{\pm 1}$ & $0.004^{\pm 0.004}$ \\ 
        $\lambda_{\text{emptiness}} = 10.0$, $\lambda_{\text{SI}} = 0.0$ & $42^{\pm 1}$ & $0.014^{\pm 0.004}$ \\ 
        $\lambda_{\text{emptiness}} = 15.0$, $\lambda_{\text{SI}} = 0.0$ & $40^{\pm 1}$ & $0.007^{\pm 0.004}$ \\ 
        $\lambda_{\text{emptiness}} = 0.0$, $\lambda_{\text{SI}} = 1.0$ & $625^{\pm 7}$ & $4.117^{\pm 0.358}$ \\ 
        $\lambda_{\text{emptiness}} = 5.0$, $\lambda_{\text{SI}} = 1.0$ & $41^{\pm 1}$ & $0.052^{\pm 0.005}$ \\ 
        $\lambda_{\text{emptiness}} = 10.0$, $\lambda_{\text{SI}} = 1.0$ & $41^{\pm 1}$ & $0.054^{\pm 0.005}$ \\ 
        $\lambda_{\text{emptiness}} = 15.0$, $\lambda_{\text{SI}} = 1.0$ & $42^{\pm 1}$ & $0.049^{\pm 0.005}$ \\ 
        $\lambda_{\text{emptiness}} = 0.0$, $\lambda_{\text{SI}} = 0.1$ & $907^{\pm 15}$ & $5.273^{\pm 0.439}$ \\ 
        $\lambda_{\text{emptiness}} = 5.0$, $\lambda_{\text{SI}} = 0.1$ & $39^{\pm 1}$ & $0.029^{\pm 0.004}$ \\ 
        $\lambda_{\text{emptiness}} = 10.0$, $\lambda_{\text{SI}} = 0.1$ & $42^{\pm 1}$ & $0.008^{\pm 0.005}$\\ 
        $\lambda_{\text{emptiness}} = 15.0$, $\lambda_{\text{SI}} = 0.1$ & $39^{\pm 1}$ & $0.021^{\pm 0.003}$ \\ 
        $\lambda_{\text{emptiness}} = 0.0$, $\lambda_{\text{SI}} = 0.01$ & $1259^{\pm 9}$ & $0.636^{\pm 0.094}$ \\ 
        $\lambda_{\text{emptiness}} = 5.0$, $\lambda_{\text{SI}} = 0.01$ & $39^{\pm 1}$ & $0.010^{\pm 0.004}$ \\ 
        $\lambda_{\text{emptiness}} = 10.0$, $\lambda_{\text{SI}} = 0.01$ & $39^{\pm 1}$ & $0.004^{\pm 0.005}$ \\ 
        $\lambda_{\text{emptiness}} = 15.0$, $\lambda_{\text{SI}} = 0.01$ & $38^{\pm 1}$ & $0.012^{\pm 0.004}$ \\
    \bottomrule
    \end{tabular}
        }
    \caption{Hyperparameter tuning for the sphere proxy. We evaluate how well the sphere proxy approximates the surface and the volume of the corresponding mesh. $\pm$ indicates the 95\% confidence interval.}
    \label{tab:supp:hyperparameters-sp}
\end{table}
\begin{table*}[ht]
    \centering
    \resizebox{1.0\linewidth}{!}{
    \begin{tabular}{lcccccccccc}
        \toprule
        \multirow{2}{2.cm}{\centering Method} & \multirow{2}{2.cm}{\centering R Precision top 1$\uparrow$} & 
        \multirow{2}{2.cm}{\centering R Precision top 2$\uparrow$} & 
        \multirow{2}{2.cm}{\centering R Precision top 3$\uparrow$} & 
        \multirow{2}{1.5cm}{\centering FID$\downarrow$} &
        \multirow{2}{2.5cm}{\centering Multimodal Dist$\downarrow$} &
        \multirow{2}{2cm}{\centering Diversity$\rightarrow$} &
        \multirow{2}{2cm}{\centering Multimodality$\uparrow$} &
        \multirow{2}{2cm}{\centering SI$\downarrow$} &
        \multirow{2}{2.cm}{\centering Surface $\downarrow$} &
        \multirow{2}{2.cm}{\centering VolDev $\downarrow$} \\
        \\
        \midrule
        128 spheres, 50\% detail & $0.440^{\pm 0.007}$ & $0.633^{\pm 0.006}$ & $0.736^{\pm 0.006}$ & $0.411^{\pm 0.056}$ & $3.450^{\pm 0.024}$ & $9.689^{\pm 0.107}$ & $2.711^{\pm 0.059}$ & $306^{\pm 11}$ & $43^{\pm 1}$ & $0.029^{\pm 0.004}$ \\
        128 spheres, 75\% detail & $0.458^{\pm 0.006}$ & $0.649^{\pm 0.006}$ & $0.753^{\pm 0.006}$ & $0.278^{\pm 0.023}$ & $3.311^{\pm 0.028}$ & $9.752^{\pm 0.078}$ & $2.583^{\pm 0.056}$ & $336^{\pm 09}$ & $44^{\pm 1}$ & $0.039^{\pm 0.005}$ \\
        192 spheres, 50\% detail & $0.435^{\pm 0.005}$ & $0.628^{\pm 0.006}$ & $0.731^{\pm 0.006}$ & $0.265^{\pm 0.024}$ & $3.462^{\pm 0.026}$ & $9.568^{\pm 0.086}$ & $2.893^{\pm 0.075}$ & $382^{\pm 09}$ & $42^{\pm 1}$ & $0.008^{\pm 0.005}$ \\
        192 spheres, 75\% detail & $0.432^{\pm 0.007}$ & $0.622^{\pm 0.006}$ & $0.725^{\pm 0.007}$ & $0.292^{\pm 0.023}$ & $3.444^{\pm 0.028}$ & $9.612^{\pm 0.050}$ & $2.762^{\pm 0.082}$ & $329^{\pm 08}$ & $37^{\pm 1}$ & $0.026^{\pm 0.004}$ \\
        256 spheres, 50\% detail & $0.429^{\pm 0.006}$ & $0.619^{\pm 0.005}$ & $0.725^{\pm 0.006}$ & $0.301^{\pm 0.038}$ & $3.437^{\pm 0.024}$ & $9.551^{\pm 0.090}$ & $2.754^{\pm 0.061}$ & $331^{\pm 05}$ & $35^{\pm 1}$ & $0.010^{\pm 0.004}$ \\
        256 spheres, 75\% detail & $0.417^{\pm 0.007}$ & $0.612^{\pm 0.006}$ & $0.718^{\pm 0.005}$ & $0.465^{\pm 0.039}$ & $3.525^{\pm 0.029}$ & $9.583^{\pm 0.085}$ & $2.782^{\pm 0.068}$ & $391^{\pm 09}$ & $32^{\pm 1}$ & $0.021^{\pm 0.004}$ \\
        \hline
    \end{tabular}
    }
    \caption{Evaluation of the number of spheres and detail samples during the training of the sphere proxy. \textit{detail} refers to the percentage of surface samples corresponding to the hand and feet regions of the original mesh. \textit{Surface} and \textit{VolDev} are evaluated on the sphere proxy test set, and the other metrics are evaluated on the HumanML3D~\cite{GZZ+22-humanml3d} test set. $\rightarrow$ indicates closer to real is better. $\pm$ indicates the 95\% confidence interval.}
    \label{tab:supp:oversampling}
\end{table*}
\paragraph{Intersection reduction}
In \cref{subsec:meth:SI-loss}, we explain that we do not use all sphere pairs to calculate our novel self-intersection loss. Specifically, we calculate the sphere intersections for all poses in the HumanML3D training set and exclude all sphere pairs that intersect in more than 90\% of the poses. To investigate the influence of this threshold, we calculate the percentage of excluded sphere pairs for varying threshold values. The resulting curve is shown in \cref{fig:supp:sphere-reduction}. 1.6\% of the sphere pairs collide in every pose, which are neighboring spheres in the sphere proxy. Excluding sphere pairs that collide in 90\% of the poses results in a reduction of sphere pairs of roughly 2\%. These are mainly the spheres near the armpits and between the legs. Intersections in these areas are frequently present in the training data because the SMPL model does not model deformations due to contact. Hence, excluding sphere pairs corresponding to these areas is justified because these self-intersections result from limitations of the SMPL model, while the motions are realistic.

\begin{figure*}[ht]
    \centering
    \includegraphics[width=0.9\linewidth]{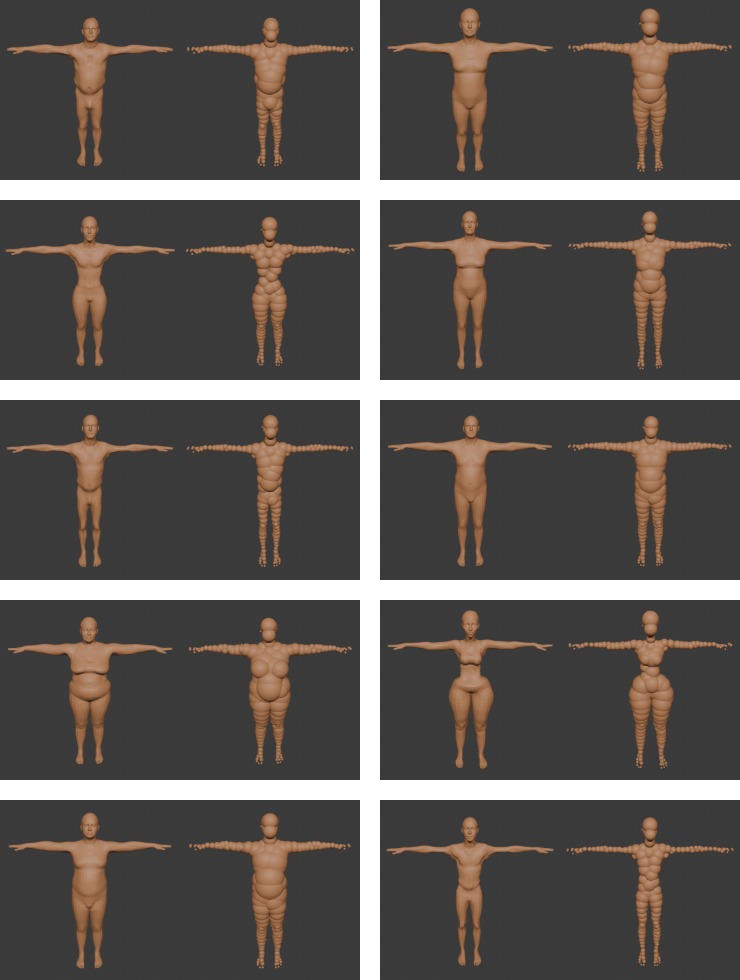}
    \caption{Visual comparison between SMPL meshes and the corresponding sphere proxies.}
    \label{fig:supp:comp_sp_smpl}
\end{figure*}

\paragraph{Number of spheres}
\cref{tab:supp:oversampling} provides additional results for the ablation study on the number of spheres. We train SIA-MDM with sphere proxies with 128, 192, and 256 spheres. During training, for each batch, we sample 10\% of SDF samples per mesh in a sphere around the mesh, while the other 90\% of SDF points are sampled close to the surface of the mesh. We explain in the main paper that we sample 50\% of the surface samples from the hand and feet regions when training the sphere proxy, which we call \textit{detail} samples. We evaluate this design choice in conjunction with the number of spheres and train SIA-MDM with sphere proxies using 50\% and 75\% detail samples.
We evaluate the sphere proxies using the proposed metrics. Looking at \textit{Surface} and \textit{VolDev}, the results are as expected - using more spheres results in a better approximation of human geometry. However, this trend is not visible in the training of SIA-MDM. Further research is necessary to determine the best settings for the sphere proxy to maximize human motion generation performance.
\paragraph{Boneweight matrix}
The spheres of the sphere proxy are attached to the underlying skeleton structure using the bone weight matrix. As described in \cref{subsec:meth:SP}, we calculate individual bone weight matrices for all training samples and use the mean of them to obtain one bone weight matrix for the sphere proxy. Having one bone weight matrix for all sphere proxies eliminates the necessity to calculate a new bone weight matrix every time a new sphere proxy is obtained. To validate this design choice, we calculate the standard deviation of the bone weight matrix. The mean of all standard deviation values is 0.01, which justifies taking the mean bone weight matrix of the training data instead of calculating a separate bone weight matrix for every sphere proxy. The highest standard deviation is 0.41. However, these high values are present in spheres at the boundary between two joints, which does not result in significant qualitative differences.
To further improve the quality of the sphere proxy, it is possible to use the mean blend weight matrix as a prior and learn a neural network, which predicts deviations of the mean blend weight matrix for individual sphere proxies. We leave this improvement for future research.
\paragraph{Joint regressor}
We randomly sample 1,500 sets of SMPL shape parameters and use the corresponding joint locations as test data for our joint regressor. We calculate the Euclidean distance between the predicted and ground truth joint locations for each joint. The mean distance between predicted and ground truth joint locations is 0.17~cm, justifying the use of the joint regressor.
\paragraph{Qualitative results}
\cref{fig:supp:comp_sp_smpl} shows a visual comparison between SMPL meshes and the corresponding sphere proxies. The SMPL shape parameters $\beta$ are randomly sampled between $-3$ and $3$. The image shows a high approximation quality for various SMPL shape parameters.

\section{Qualitative Results}
\label{sec:supp:qualitative}

\begin{figure*}[ht]
    \centering
    \includegraphics[width=0.98\linewidth]{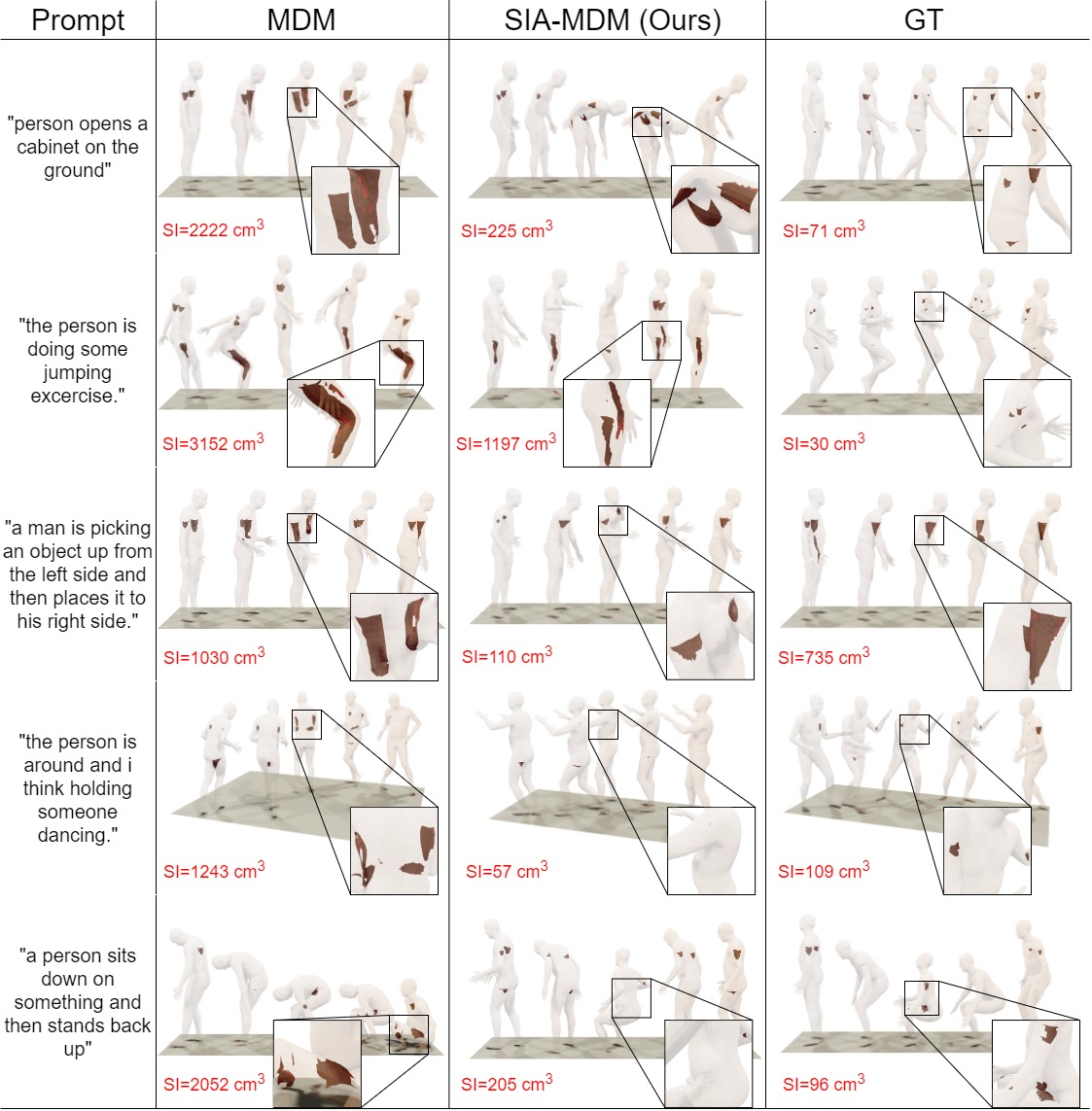}
    \caption{\textbf{Visual Comparison} between motions generated by SIA-MDM (Ours), MDM~\cite{TRG+23-MDM}, and ground truth motions (GT) given text prompts from the HumanML3D~\cite{GZZ+22-humanml3d} test set. Darker colors indicate later frames in the motion, while red patches indicate self-intersections. \textcolor{red}{SI} indicates the mean self-intersection volume of the generated motion. SIA-MDM generates significantly fewer self-intersections while semantically following the textual description. Interestingly, even the ground truth trajectories contain some self-intersections, even though they are recorded with a motion capture setup.}
    \label{fig:supp:more_qual_res}
\end{figure*}

\cref{fig:supp:more_qual_res} shows more generated motions of MDM~\cite{TRG+23-MDM} and SIA-MDM given text prompts of the HumanML3D~\cite{GZZ+22-humanml3d} test set. In addition, we show the ground truth motions. SIA-MDM generates significantly fewer self-intersections than MDM, while the motions semantically follow the textual description. The ground truth motions also contain self-intersections, mainly near the armpits and between the legs, due to the limitations of the SMPL model. The high amount of self-intersections in the third row for the ground truth motions is most likely due to the processing of HumanML3D, which maps all motions to one target skeleton. This motion was most likely recorded by a skinnier model than the target motion model. Therefore, applying the same joint rotations to the target skeleton results in many self-intersections, as shown in this image.

Furthermore, we provide videos as part of this supplementary material, comparing motions generated by our self-intersection-aware methods to their respective baselines given text prompts of the test sets of HumanML3D and KIT-ML. The videos also provide the mean self-intersections per motion using our novel \textit{SI} metric. We have fewer motions than the human motion generation evaluation, so we use smaller voxels with edge length $0.02$~cm.

\FloatBarrier

\end{document}